\documentclass[letterpaper]{article} 
\usepackage{aaai2027}  
\usepackage[hyphens]{url}  
\usepackage{graphicx} 
\usepackage{natbib}  
\usepackage{caption} 
\usepackage{booktabs}
\usepackage{amsmath,amssymb,bm}
\usepackage{multirow}
\usepackage{enumitem}
\usepackage{xcolor}
\usepackage{array}
\usepackage{longtable}
\usepackage{placeins}
\usepackage{float}
\graphicspath{{figure/}{NCGR_supplementary/figures/}}

\setlist[itemize]{leftmargin=2em}
\setlist[enumerate]{leftmargin=2em}

\title{NCGR: Noise-Conditional Gated Rectification for\\
Camera Extrinsic Perturbations in BEV 3D Object Detection}
\author{
Wenbin Pan\equalcontrib,
Wanhao Liu\equalcontrib,
Liwei Luo,
Panshuo Li\textsuperscript{\textdagger},
Yong Xu,
Renquan Lu
}
\affiliations{
Guangdong University of Technology, Guangzhou, China
}

\nocopyright

\begin{document}

\maketitle
\begingroup
\renewcommand{\thefootnote}{}
\footnotetext{\textsuperscript{\textdagger}Corresponding author.}
\addtocounter{footnote}{-1}
\endgroup

\begin{abstract}
Camera-based bird's-eye-view (BEV) 3D detection typically assumes accurate and fixed camera extrinsics. In detectors using spatial cross-attention (SCA), extrinsic perturbations displace the image-plane projections of BEV reference points, causing queries to sample features from incorrect regions and degrading detection performance. To address this failure mode, Noise-Conditional Gated Rectification (NCGR) is proposed to compensate for projection errors without explicitly estimating a full six-degree-of-freedom extrinsic correction. For each query--camera pair, a 2D rectification offset is predicted and modulated by a camera-level gate to rectify the base projection before native deformable sampling. During training, the perturbation-derived quantities used to construct the condition and gate are gradually replaced through scheduled interpolation by counterparts generated from an auxiliary scalar predicted from camera features. This transition enables blind inference without perturbation metadata. During training, a weight-shared clean-teacher/perturbed-student pair is used, and the rectification module is supervised by a BEV-consistency objective between the two branches. NCGR is evaluated on nuScenes with simulated dynamic and static extrinsic perturbations. In a five-camera dynamic stress test, NCGR achieves \(39.69\%\) NDS, compared with \(28.00\%\) for BEVFormer and \(33.23\%\) for CAPE. Under clean extrinsics, NCGR maintains performance comparable to that of BEVFormer.
\end{abstract}

\section{Introduction}

In multi-camera 3D object detection, a bird's-eye-view (BEV) representation is commonly constructed through either depth-based lifting~\cite{lss,bevdet,bevdepth} or attention-based interaction between 3D/BEV queries and image features~\cite{detr3d,petr,bevformer}. Because depth-based lifting depends on estimated depth distributions, errors in these distributions can be propagated into BEV construction. Attention-based methods instead support geometry-guided multi-view aggregation without explicitly predicting dense per-pixel depth. A representative design is BEVFormer's spatial cross-attention (SCA), which establishes image--BEV correspondence by explicitly projecting 3D reference points into camera views. Its sampling locations are therefore directly dependent on camera extrinsics.

\begin{figure}[t]
    \centering
    \includegraphics[width=\linewidth]{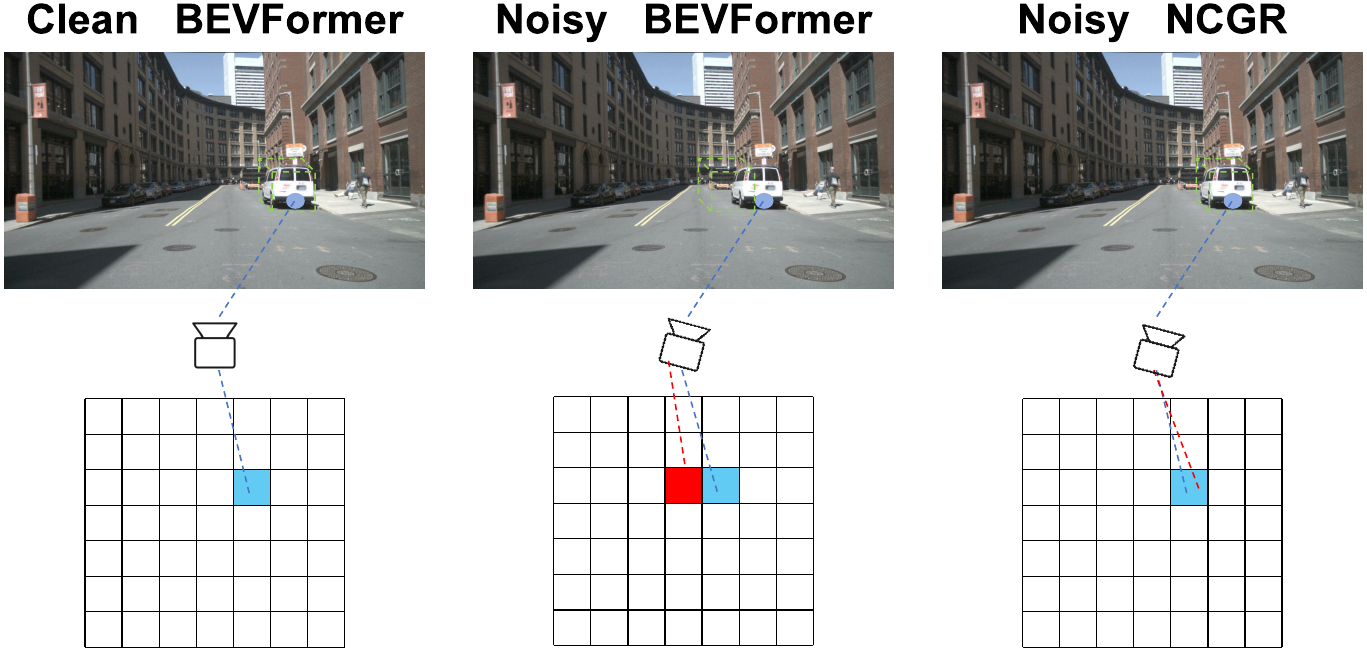}
    \caption{Motivation and problem illustration. Camera extrinsic perturbations shift the image-plane projections of BEV reference points, causing spatial cross-attention to sample misaligned image features and degrading BEV detection.}
    \label{fig:motivation}
\end{figure}

This geometric dependence creates a specific failure mode under extrinsic perturbations. Perturbed extrinsics displace the base projection of each reference point, causing the corresponding BEV query to aggregate features from mismatched image regions, as illustrated in Fig.~\ref{fig:motivation}. Native multi-scale deformable attention is not designed to explicitly correct this anchor displacement. The module uses its learned offsets to sample locally around the projected anchor, which remains misaligned under extrinsic drift~\cite{deformable_detr,bevformer}. 

Existing approaches address calibration errors differently. Explicit recovery methods estimate corrected camera--LiDAR transformations, but generally require cross-modal observations or a separate calibration stage~\cite{calibnet,lccnet,calibformer,bevcalib}. Detector-side methods improve tolerance through augmentation, positional encoding, view transformation, or enlarged sampling neighborhoods~\cite{lss,bevformer,petrv2,widthformer,garkt}, while calibration-free or camera-local representations reduce reliance on precise extrinsics~\cite{cft,cbr,cape,graphbev}. Explicit recovery operates outside the detector. Tolerance-oriented methods generally do not explicitly update the displaced projection anchor. Calibration-free and camera-local designs instead alter how explicit geometry is incorporated. Directly rectifying the perturbed base projection inside SCA while retaining SCA geometry remains underexplored.

To address the projection displacement caused by extrinsic perturbations, Noise-Conditional Gated Rectification (NCGR) is proposed. NCGR applies a gated, query--camera-specific 2D residual offset to each displaced base projection before native deformable sampling. Unlike explicit recovery, NCGR retains the original geometric projection as the reference and applies a residual correction without explicitly estimating a full six-degree-of-freedom extrinsic correction. In a five-camera dynamic stress test on nuScenes, NCGR achieves \(39.69\%\) NDS, compared with \(28.00\%\) for BEVFormer and \(33.23\%\) for CAPE, while maintaining comparable accuracy under clean extrinsics.

The main contributions are:
\begin{itemize}
    \item The analysis distinguishes the displacement of SCA base projections from the local offsets used by native deformable attention.
    \item NCGR introduces gated residual rectification inside SCA, retains the explicit projection as the geometric reference, and transitions from perturbation-derived controls to a learned camera-level signal for inference.
    \item Experiments demonstrate improved robustness over BEVFormer and CAPE under severe multi-camera perturbations, with comparable performance maintained under clean extrinsics.
\end{itemize}

\section{Related Work}

\subsection{Explicit Extrinsic Recovery}

Explicit extrinsic-recovery methods estimate the rigid transformation between cameras and LiDAR from cross-modal observations. RegNet, CalibNet, and LCCNet estimate six-degree-of-freedom extrinsic corrections through deep cross-modal registration, geometric and photometric consistency, and cost-volume-based iterative refinement, respectively~\cite{regnet,calibnet,lccnet}. SemAlign optimizes extrinsics through annotation-free semantic alignment~\cite{semalign}, while CalibFormer and LCCRAFT improve calibration accuracy using multi-head cross-modal correlation modeling and recurrent updates over all-pairs correlations, respectively~\cite{calibformer,lccraft}. For downstream perception, the recalibration method of Song et al.\ predicts extrinsic deviations through image--point-cloud semantic alignment, whereas BEVCalib decodes global extrinsic corrections from shared BEV features~\cite{recalibration,bevcalib}. These methods directly optimize sensor-level rigid transformations and return corrected extrinsics to downstream modules. However, they generally depend on LiDAR or other cross-modal observations and act as a calibration stage separate from detection: the recovered extrinsics are fed back into the projection pipeline rather than being used to correct feature sampling inside the detector. Any remaining error in the recovered extrinsics can therefore limit downstream detection performance.

\subsection{Detector-Side Robustness to Extrinsic Perturbations}

One family of detector-side methods retains the calibrated projection while making the detector more tolerant to extrinsic errors through training augmentation, architectural design, or BEV denoising~\cite{robust3d,viewpoint_robust,robobev,bevdiffuser,resbev}. Lift-Splat-Shoot applies extrinsic perturbations and camera dropout during training~\cite{lss}, and BEVFormer uses spatial cross-attention and temporal self-attention to aggregate multi-view image features and historical BEV information, respectively~\cite{bevformer}. PETRv2 improves robustness through its positional encoding design~\cite{petrv2}. WidthFormer evaluates extrinsic robustness in the context of its view-transformation design~\cite{widthformer}. GARKT enlarges the sampling kernels around coarse projection locations to address displacement in tire-blow-out scenarios. The kernel size is selected according to predefined impact levels~\cite{garkt}. These methods improve tolerance through the training distribution,
positional representations, or enlarged sampling neighborhoods, but they do
not explicitly update the displaced base projection. Approaches that adjust
the sampling range may additionally rely on predefined perturbation levels.

Another family of detector-side methods reduces reliance on precise calibration by changing how camera geometry is incorporated into representation construction and feature fusion~\cite{monocular_free}. CFT and the roadside-oriented CBR avoid camera calibration parameters and learn image-to-BEV mappings implicitly~\cite{cft,cbr}, whereas CAPE and SpatialDETR organize cross-view interaction with camera-local coordinate representations and camera-ray encodings, respectively~\cite{cape,spatialdetr}. In multimodal perception, TransFusion relaxes hard projection correspondences through soft attention-based association, and GraphBEV aligns features during depth projection and BEV fusion~\cite{transfusion,autoalign,bevdistill,graphbev}. These designs reduce sensitivity to calibration errors. However, they either use geometric representations other than the explicit projection employed by SCA or focus on roadside and multimodal settings rather than the dynamic per-camera extrinsic drift considered here.

Existing approaches recover extrinsics, improve tolerance without updating displaced projection anchors, or alter how explicit geometry is incorporated. Directly rectifying displaced base projections inside SCA remains underexplored, especially when perturbation metadata are unavailable at inference. NCGR addresses this gap by retaining the original geometric projection as the reference and applying a residual correction.

\begin{figure*}[t]
    \centering
    \includegraphics[width=1\textwidth]{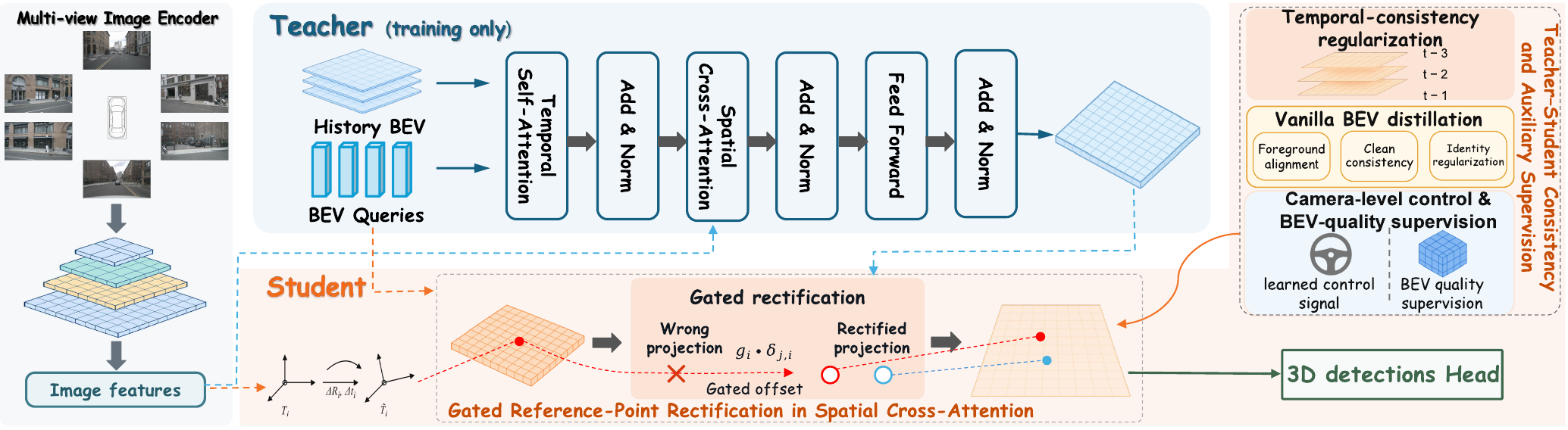}
    \caption{Overall framework of NCGR. The student branch uses perturbed extrinsics and gated reference-point rectification, while the teacher branch provides clean BEV targets. During training, synthetic perturbations provide the signals used to form the correction condition and gate according to the interpolation schedule. These perturbation-derived targets are used only for training. During inference, a learned camera-level scalar provides a bounded,
metadata-free control signal for the correction condition and gate.}
    \label{fig:framework}
\end{figure*}

\section{Method}

\paragraph{Problem formulation.}
In SCA, a 3D reference point $\mathbf p$ is projected into the image plane of
camera $i$ by the calibrated LiDAR-to-image projection matrix
$\mathbf T_{\mathrm{l2i}}^{(i)}$, yielding the base location
$\boldsymbol{\pi}_i(\mathbf p)$. Let
$\mathbf T_{\mathrm{l2c}}^{(i)}$ denote the LiDAR-to-camera extrinsic
transformation of camera $i$. During training, each selected camera is
assigned a synthetic rigid perturbation $\Delta\mathbf T_i$, whose rotation is
parameterized by the Euler-angle vector
$\boldsymbol{\alpha}_i=[\phi_i,\vartheta_i,\psi_i]^\top$ and whose translation
is $\Delta\mathbf t_i$. The perturbed extrinsic transformation and the resulting image-plane
displacement are
\begin{equation}
\begin{aligned}
\widetilde{\mathbf T}_{\mathrm{l2c}}^{(i)}
&=
\Delta\mathbf T_i\mathbf T_{\mathrm{l2c}}^{(i)},\\
\Delta\boldsymbol{\pi}_i(\mathbf p)
&=
\widetilde{\boldsymbol{\pi}}_i(\mathbf p)
-
\boldsymbol{\pi}_i(\mathbf p),
\end{aligned}
\label{eq:compact_perturbation}
\end{equation}
where $\widetilde{\boldsymbol{\pi}}_i(\mathbf p)$ is obtained by combining the
perturbed extrinsic transformation with the camera intrinsics. This displacement moves the SCA
base projection to an incorrect image location, causing the corresponding BEV
query to sample misaligned image features. NCGR therefore rectifies the
displaced base projection before native deformable sampling. The complete
matrix formulation, coordinate convention, and perturbation protocol are
provided in the supplementary material.

\subsection{Method Overview}

As illustrated in Fig.~\ref{fig:framework}, NCGR constructs
clean--perturbed training pairs. The framework contains two main components:
\begin{enumerate}[leftmargin=2em]
    \item \textbf{Gated reference-point rectification in spatial cross-attention}, which
    applies gated 2D residual offsets to the projected reference
    points inside SCA to mitigate sampling misalignment caused by perturbed
    extrinsics;
    \item \textbf{BEV consistency between the clean teacher and perturbed
    student branches}, which constrains the perturbed student BEV representation
    using the corresponding clean teacher representation, thereby stabilizing
    training under extrinsic perturbations and preserving performance under clean extrinsics.
\end{enumerate}

The BEV-consistency objective is implemented through two forward passes with
shared weights during training:
\begin{itemize}[leftmargin=2em]
    \item \textbf{Student path}: uses the perturbed LiDAR-to-image projection
    matrices and enables the rectification modules;
    \item \textbf{Teacher path}: uses the unperturbed projection matrices,
    disables the rectification modules, and is run without gradient
    computation.
\end{itemize}

For inference, an auxiliary camera-level head maps each camera feature to a
bounded scalar that controls both the correction condition and the multiplicative
gate. Scheduled interpolation replaces perturbation-derived training controls
with this learned signal. The perturbation-derived targets are used only during
training. Because perturbations modify projection metadata rather than image
appearance, the scalar provides a bounded operating point shaped by the training
distribution, while the query--camera offset network predicts the spatial
adjustment. Training
also uses camera-level control supervision, BEV-level auxiliary supervision,
and a low-weight temporal-consistency regularizer.

\subsection{Gated Reference-Point Rectification in Spatial Cross-Attention}
\label{sec:ncsca}

\paragraph{Training-time control construction and scheduling.}
For each camera $i$,
the synthetic perturbation is summarized by
\begin{equation}
\mathbf n_i
=
\left[
n_i^{\mathrm{rot}},
n_i^{\mathrm{trans}}
\right]^\top
=
\left[
\|\boldsymbol{\alpha}_i\|_2,
\|\Delta\mathbf t_i\|_2
\right]^\top
\in\mathbb{R}^2,
\label{eq:mag}
\end{equation}
where $n_i^{\mathrm{rot}}$ and $n_i^{\mathrm{trans}}$ denote the rotation and
translation perturbation magnitudes, respectively, and
$\mathbf n_i=\mathbf{0}_2$ for an unperturbed camera. After normalization by the per-axis rotation and translation bounds
$\sigma_r$ and $\sigma_t$, respectively, one obtains
\begin{equation}
\begin{aligned}
    \bar n_i^{\mathrm{rot}}
    &= \operatorname{clip}\!\left(
    \frac{n_i^{\mathrm{rot}}}{\sqrt{3}\,\sigma_r},0,2\right),\\
    \bar n_i^{\mathrm{trans}}
    &= \operatorname{clip}\!\left(
    \frac{n_i^{\mathrm{trans}}}{\sqrt{3}\,\sigma_t},0,2\right),
\end{aligned}
\qquad
    \bar{\mathbf n}_i=
    \left[\bar n_i^{\mathrm{rot}},\bar n_i^{\mathrm{trans}}\right]^{\top}.
    \label{eq:normalized_noise}
\end{equation}
Since $|\phi_i|,|\vartheta_i|,|\psi_i|\leq\sigma_r$, the rotation
magnitude satisfies
$
\|\boldsymbol{\alpha}_i\|_2
=
\sqrt{\phi_i^2+\vartheta_i^2+\psi_i^2}
\leq
\sqrt{3\sigma_r^2}
=
\sqrt{3}\sigma_r.
$
The same reasoning gives the translation upper bound
$\|\Delta\mathbf t_i\|_2\leq\sqrt{3}\sigma_t$.

An auxiliary camera-level control head $\mathcal H$ applies global
spatial pooling followed by a two-layer multilayer perceptron (MLP) with a
sigmoid output. For camera $i$, it predicts
$
    \widehat q_i=\mathcal H(\mathbf F_i)\in[0,1],
$
where $\mathbf F_i$ is the top-level feature of camera $i$. During training,
the normalized rotation and translation perturbation magnitudes are combined
to define
$
    g_i^{\mathrm{gt}}
    =
    \operatorname{clip}\!\left(
    \max\!\left(\bar n_i^{\mathrm{rot}},\bar n_i^{\mathrm{trans}}\right),
    0,1
    \right),
$
and
$
    q_i^\star=1-g_i^{\mathrm{gt}}.
$
Here, $g_i^{\mathrm{gt}}$ denotes the perturbation-derived gate target, while
its complement $q_i^\star$ denotes the camera-health target, which equals one
for an unperturbed camera and decreases as the perturbation magnitude increases. The learned scalar is converted to the
camera-level control
$
    \rho_i=1-\operatorname{sg}(\widehat q_i),
$
where $\operatorname{sg}(\cdot)$ denotes stop-gradient. The camera-level
control head is supervised by
\begin{equation}
    \mathcal L_{\mathrm{health}}^{\mathrm{cam}}
    =
    \frac{1}{N}
    \sum_{i=1}^{N}
    \left(\widehat q_i-q_i^\star\right)^2,
    \label{eq:gate_parameterization}
\end{equation}
where $N$ is the number of cameras. The loss is averaged over all camera
instances in the mini-batch. The scalar $\rho_i$ is used to construct both
the correction condition and the multiplicative gate. During training, they
are scheduled as
\begin{equation}
\begin{aligned}
    \mathbf c_i
    &= (1-\alpha)\bar{\mathbf n}_i+\alpha\rho_i\mathbf{1}_2,\\
    g_i
    &= (1-\alpha)g_i^{\mathrm{gt}}+\alpha\rho_i,
    \qquad g_i\in[0,1],
\end{aligned}
\label{eq:condition_gate_schedule}
\end{equation}
where $\mathbf{1}_2=[1,1]^\top$. Let $s=e/E$ denote the training progress,
where $e$ is the current epoch and $E$ is the total number of epochs. The
scheduling coefficient is
\begin{equation}
    \alpha
    =
    \operatorname{clip}\!\left(
    \frac{s-0.3}{0.4},
    0,1
    \right).
    \label{eq:gate_alpha}
\end{equation}
At inference, $\alpha$ is fixed at its terminal value of $1$. Thus,
$
    \mathbf c_i=\rho_i[1,1]^\top, g_i=\rho_i,
$
and both quantities are derived solely from the learned camera-level signal
without perturbation metadata.

\paragraph{Gated reference-point rectification.}
Let $N_b$ denote the mini-batch size. The symbols $N_q$ and $N$ denote the
numbers of BEV queries and cameras, respectively. The encoder contains $L$ SCA
layers, and each BEV reference point has $D$ height anchors.

The index $j\in\{1,\ldots,N_q\}$ identifies a BEV query, and
$i\in\{1,\ldots,N\}$ identifies a camera. The index
$d\in\{1,\ldots,D\}$ identifies a height anchor; the current implementation
uses $D=4$.

For clarity, Eqs.~\eqref{eq:raw_offset}--\eqref{eq:final_sampling} describe
one sample and one SCA layer. The mini-batch index
$b\in\{1,\ldots,N_b\}$ and SCA-layer index
$\ell\in\{1,\ldots,L\}$ are omitted. With these indices restored,
$\boldsymbol{\delta}_{j,i}$ is
$\boldsymbol{\delta}_{\ell,b,j,i}$, while $\mathbf c_i$ and $g_i$ are 
$\mathbf c_{b,i}$ and $g_{b,i}$, respectively.

For the $j$-th BEV query and the $i$-th camera, let
$\mathcal G^{(i)}(\cdot)\in\mathbb R^2$ denote the $i$-th 2D output slice
of the shared correction network $\mathcal G$. The bounded image-plane offset is
\begin{equation}
    \boldsymbol{\delta}_{j,i}
    = s_{\delta}\tanh\!\left(
    \mathcal{G}^{(i)}\!\left([\mathbf Q_j;\mathbf c_i]\right)
    \right),
    \qquad \boldsymbol{\delta}_{j,i}\in\mathbb{R}^{2},
    \label{eq:raw_offset}
\end{equation}
where $\mathbf Q_j\in\mathbb R^{256}$ denotes the $j$-th BEV query,
$\mathbf c_i$ is the camera-level condition for camera $i$, and $s_{\delta}$
denotes the maximum absolute rectification offset in the normalized
image-plane coordinates used by SCA. In each SCA
layer, $\mathcal G$ consists of a $258$-to-$128$ linear layer, a ReLU
activation, and a $128$-to-$12$ output linear layer. The output
$\mathcal G([\mathbf Q_j;\mathbf c_i])\in\mathbb R^{2N}$ packs one 2D offset
for each camera; with $N=6$, its dimension is 12. The slice
$\mathcal G^{(i)}([\mathbf Q_j;\mathbf c_i])$ provides the residual sampling
offset for query $j$ and camera $i$. The same $\mathcal G$ is shared across all BEV
queries and cameras within one SCA layer, while the six encoder layers use
separate correction networks.
The offset is expressed in the normalized image-plane coordinates used by
SCA. The scaled hyperbolic tangent function bounds each offset component to
$[-s_{\delta},s_{\delta}]$. The first
linear layer of $\mathcal G$ uses Xavier-uniform weight initialization with zero
bias, while its output linear layer is initialized to zero. The six correction
networks introduce approximately $0.208$M parameters. Practical runtime and
memory overheads are reported in the supplementary material.

The effective offset is
\begin{equation}
    \boldsymbol{\delta}_{j,i}^{\star}
    = g_i\boldsymbol{\delta}_{j,i}.
    \label{eq:eff_offset}
\end{equation}
Let $\mathbf r_{j,i,d}\in\mathbb{R}^{2}$ denote the normalized image-plane
reference point of the $d$-th height anchor associated with query $j$ in camera
$i$. Spatial cross-attention uses the rectified location
\begin{equation}
    \mathbf r_{j,i,d}^{\mathrm{rect}}
    = \mathbf r_{j,i,d}
    + \boldsymbol{\delta}_{j,i}^{\star},
    \label{eq:rectified_reference}
\end{equation}
where the same query--camera offset is applied to all height anchors of the
projected BEV reference point. The native deformable-attention module remains unchanged and predicts local
sampling offsets from the query, which are applied around each rectified
anchor. Let
$\Delta\mathbf r_{j,i,d,h,m}^{\mathrm{nat}}\in\mathbb{R}^{2}$ denote the
native sampling offset after normalization by the spatial size of the associated
feature map, where $h$ and $m$ index the attention head
and sampling point, respectively. The final sampling location is
\begin{equation}
\mathbf s_{j,i,d,h,m}
=
\mathbf r_{j,i,d}^{\mathrm{rect}}
+
\Delta\mathbf r_{j,i,d,h,m}^{\mathrm{nat}}.
\label{eq:final_sampling}
\end{equation}
NCGR first shifts the geometrically projected base reference point by
$\boldsymbol{\delta}_{j,i}^{\star}$. The native deformable-attention module
then uses $\Delta\mathbf r_{j,i,d,h,m}^{\mathrm{nat}}$ to sample local,
content-adaptive features around the rectified location.

\begin{figure}[t]
    \centering
    \includegraphics[width=\linewidth]{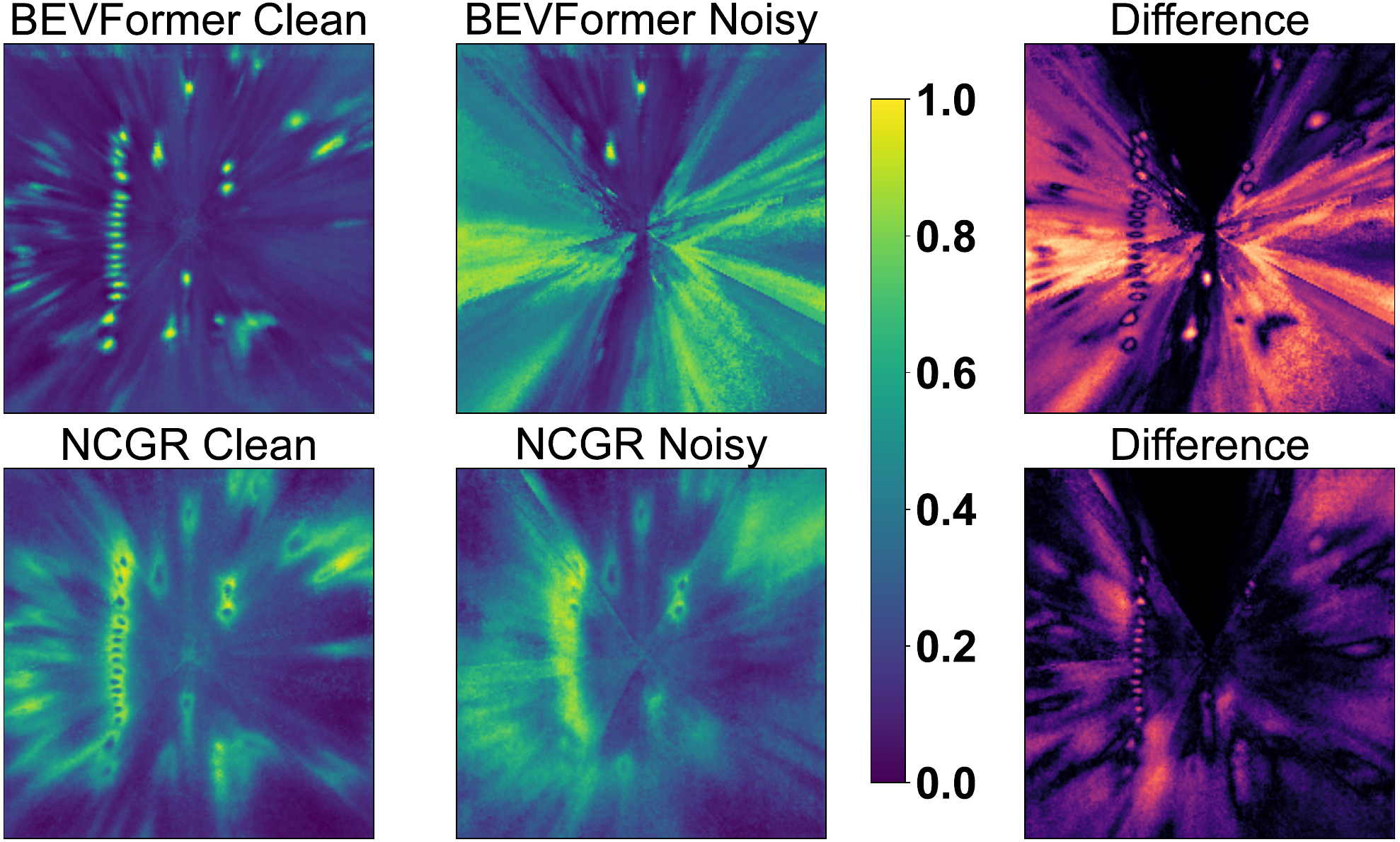}
    \caption{Qualitative comparison of clean and perturbed BEV features from BEVFormer and NCGR, together with the corresponding difference maps.}
    \label{fig:bev_heatmap}
\end{figure}

Equations~\eqref{eq:condition_gate_schedule}--\eqref{eq:final_sampling}
define the complete rectification process.
The detection and BEV-consistency losses update $\mathcal G$ through the
rectified student BEV. The stop-gradient operation blocks these losses from
updating $\mathcal H$. The camera-level control head $\mathcal H$ is trained only with
$\mathcal L_{\mathrm{health}}^{\mathrm{cam}}$ in
Eq.~\eqref{eq:gate_parameterization}.
Figure~\ref{fig:bev_heatmap} compares clean and perturbed BEV features
together with their difference maps. Relative to BEVFormer, NCGR retains a
spatial response pattern closer to its clean counterpart and exhibits weaker
perturbation-induced discrepancies, suggesting that reference-point
rectification reduces BEV feature distortion under extrinsic perturbations.

\subsection{Teacher--Student Consistency and Auxiliary Supervision}

Let
$\mathcal T=\{\mathbf T_{\mathrm{l2i}}^{(i)}\}_{i=1}^{N}$ and
$\widetilde{\mathcal T}
=\{\widetilde{\mathbf T}_{\mathrm{l2i}}^{(i)}\}_{i=1}^{N}$
denote the sets of unperturbed and perturbed LiDAR-to-image projection
matrices, respectively. Using the rectified reference points in
Eq.~\eqref{eq:rectified_reference}, the perturbed student branch produces
\begin{equation}
    \mathbf B^{\mathrm{s}}
    = \mathcal{E}\!\left(
    \mathbf F,\widetilde{\mathcal T};
    \mathcal R^{\mathrm{rect}}\right),
    \label{eq:student_bev}
\end{equation}
where $\mathcal E$ is the BEV encoder with NCGR enabled, and $\mathbf F$
contains the multi-view image features shared by both branches. The sets
$\mathcal R^{\mathrm{rect}}=\{\mathbf r_{j,i,d}^{\mathrm{rect}}\}$ and
$\mathcal R=\{\mathbf r_{j,i,d}\}$ contain the rectified and original
reference points, respectively. The clean teacher branch uses shared weights, accurate extrinsics, no
rectification, and the original reference points:
\begin{equation}
    \mathbf B^{\mathrm{t}}
    = \operatorname{sg}\!\left[
    \mathcal{E}\!\left(
    \mathbf F,\mathcal T;\mathcal R\right)
    \right].
    \label{eq:teacher_bev}
\end{equation}
The student and teacher tensors satisfy
$\mathbf B^{\mathrm{s}},\mathbf B^{\mathrm{t}}
\in\mathbb{R}^{N_b\times|\Omega|\times C}$, where $\Omega$ is the set of
BEV cells, $C$ is the feature dimension, and
$\mathbf B_{b,x}^{\mathrm{s}},\mathbf B_{b,x}^{\mathrm{t}}\in\mathbb R^C$
denote the student and teacher feature vectors at BEV cell $x\in\Omega$ of
sample $b\in\{1,\ldots,N_b\}$, respectively.

\paragraph{Vanilla BEV distillation.}
Let $M_{b,x}\in\{0,1\}$ denote the foreground mask rendered from the
ground-truth 3D boxes at BEV cell $x$ of sample $b$. The normalized foreground weight is
\begin{equation}
    w_{b,x}
    = \frac{M_{b,x}+\beta}
    {\frac{1}{|\Omega|}\sum_{y\in\Omega}(M_{b,y}+\beta)},
    \qquad \beta=0.1.
    \label{eq:foreground_weight}
\end{equation}
The foreground-weighted student--teacher alignment loss is
\begin{equation}
    \mathcal L_{\mathrm{align}}
    = \frac{1}{N_b|\Omega|}
    \sum_{b=1}^{N_b}\sum_{x\in\Omega}
    w_{b,x}\left[1-\operatorname{cos}\!\left(
    \mathbf B_{b,x}^{\mathrm{s}},
    \mathbf B_{b,x}^{\mathrm{t}}\right)\right],
    \label{eq:align_loss}
\end{equation}
where $\operatorname{cos}(\cdot,\cdot)$ denotes cosine similarity.
Foreground weighting reduces the influence of the static background, which
occupies most of the BEV map. Let $\mathcal C$ denote the set of fully unperturbed
samples in the current mini-batch. The whole-map consistency loss for these samples is
\begingroup
\small
\begin{equation}
\mathcal L_{\mathrm{clean}}
=
\begin{cases}
\displaystyle
\frac{
\sum_{b\in\mathcal C}\sum_{x\in\Omega}
\left[1-\operatorname{cos}\!\left(
\mathbf B_{b,x}^{\mathrm{s}},
\mathbf B_{b,x}^{\mathrm{t}}\right)\right]
}{
|\mathcal C||\Omega|
},
& |\mathcal C|>0,\\[8pt]
0,
& |\mathcal C|=0.
\end{cases}
\label{eq:clean_loss}
\end{equation}
\endgroup
The identity term also penalizes pre-gate offsets on unperturbed cameras. In
Eq.~\eqref{eq:identity_loss}, the previously omitted sample and layer indices
are restored. Let $\mathcal I_{\mathrm{clean}}$ denote the camera instances
that remain unperturbed in the current mini-batch, including those from samples
in which other cameras are perturbed. The identity regularizer is
\begingroup
\small
\begin{equation}
\mathcal L_{\mathrm{id}}
=
\begin{cases}
\displaystyle
\frac{
\sum_{\ell=1}^{L}
\sum_{(b,i)\in\mathcal I_{\mathrm{clean}}}
\sum_{j=1}^{N_q}
\left\|\boldsymbol{\delta}_{\ell,b,j,i}\right\|_2^2
}{
L N_q|\mathcal I_{\mathrm{clean}}|
},
& |\mathcal I_{\mathrm{clean}}|>0,\\[8pt]
0,
& |\mathcal I_{\mathrm{clean}}|=0.
\end{cases}
\label{eq:identity_loss}
\end{equation}
\endgroup
The Vanilla BEV distillation loss is therefore
\begin{equation}
    \mathcal{L}_{\mathrm{vanilla}}
    = \mathcal{L}_{\mathrm{align}}
    + \lambda_{\mathrm{clean}}\mathcal{L}_{\mathrm{clean}}
    + \lambda_{\mathrm{id}}\mathcal{L}_{\mathrm{id}},
    \label{eq:vanilla_loss}
\end{equation}
where $\lambda_{\mathrm{clean}}$ and $\lambda_{\mathrm{id}}$ are the
corresponding loss weights. Equations~\eqref{eq:align_loss} and~\eqref{eq:clean_loss} align the student
BEV with the teacher after rectified SCA sampling. Equation~\eqref{eq:identity_loss}
directly penalizes correction offsets on clean cameras.

\paragraph{Auxiliary camera-level control and BEV-quality supervision.}
The auxiliary supervision combines the camera-level control loss
$\mathcal L_{\mathrm{health}}^{\mathrm{cam}}$ in
Eq.~\eqref{eq:gate_parameterization} with a training-only BEV-quality loss.
For each BEV cell, the student--teacher cosine similarity is mapped to
$[0,1]$ and cubed to sharpen the target. The BEV-quality target
$q_{b,x}^{\star,\mathrm{bev}}$ is defined using stop-gradient as
\begin{equation}
    q_{b,x}^{\star,\mathrm{bev}}
    = \operatorname{sg}\!\left[
    \left(
    \frac{1+\operatorname{cos}(\mathbf B_{b,x}^{\mathrm{s}},
    \mathbf B_{b,x}^{\mathrm{t}})}{2}
    \right)^3
    \right].
    \label{eq:bev_health_target}
\end{equation}
The BEV-quality prediction is
$\widehat q_{b,x}^{\mathrm{bev}}
=\sigma(\mathcal H_{\mathrm{bev}}(\mathbf B^{\mathrm{s}})_{b,x})$, where
$\mathcal H_{\mathrm{bev}}$ is a BEV-level auxiliary head and
$\sigma(\cdot)$ is the sigmoid function. The BEV-quality loss $\mathcal L_{\mathrm{health}}^{\mathrm{bev}}$ is defined as
\begin{equation}
    \mathcal L_{\mathrm{health}}^{\mathrm{bev}}
    = \frac{1}{N_b|\Omega|}
    \sum_{b=1}^{N_b}\sum_{x\in\Omega}
    \left(\widehat q_{b,x}^{\mathrm{bev}}
    -q_{b,x}^{\star,\mathrm{bev}}\right)^2.
    \label{eq:bev_health_loss}
\end{equation}
Gradients through the prediction branch update the BEV-quality head and the
student representation. The auxiliary-supervision loss $\mathcal L_{\mathrm{health}}$ combines the
camera-level control and BEV-quality losses as
\begin{equation}
    \mathcal L_{\mathrm{health}}
    = \mathcal L_{\mathrm{health}}^{\mathrm{bev}}
    + \lambda_{\mathrm{cam}}
    \mathcal L_{\mathrm{health}}^{\mathrm{cam}},
    \label{eq:health_loss}
\end{equation}
where $\lambda_{\mathrm{cam}}$ weights the camera-level control loss.
The BEV-level branch is used only during training.

\paragraph{Auxiliary temporal-consistency regularization.}
The gated reference-point rectification in
Eqs.~\eqref{eq:raw_offset}--\eqref{eq:rectified_reference} addresses the
sampling misalignment caused by extrinsic perturbations. A separate
training-only loss $\mathcal L_{\mathrm{judge}}$, defined in
Eq.~\eqref{eq:judge_loss}, enforces consistency among historical student BEV
features.

For sample $b$, $\mathcal K_b$ is the set of valid historical candidates;
unavailable entries and the current frame are excluded. Historical student BEV
features are aligned to the current coordinate frame using the corresponding
pose transformations. For each $k\in\mathcal K_b$,
$\mathcal W_{b,k}(\mathbf B_{b,k}^{\mathrm{s}})$ denotes the aligned
historical student BEV. $\mathbf B_b^{0}$ denotes the most recent valid
historical student BEV after the same alignment. The head
$\mathcal H_{\mathrm{bev}}$ also predicts a quality map
for each historical candidate. The same pose transformation used for the
corresponding historical BEV feature is applied to this map. The spatial mean
of the aligned quality map is denoted by $\bar q_{b,k}$.
The scorer input is
\begin{equation}
\begin{aligned}
\mathbf x_{b,k}=\big[&
\operatorname{GAP}(\mathcal W_{b,k}(\mathbf B_{b,k}^{\mathrm{s}}));\,
\operatorname{GAP}(\mathbf B_b^{0});\\
&\operatorname{GAP}(
|\mathcal W_{b,k}(\mathbf B_{b,k}^{\mathrm{s}})-\mathbf B_b^{0}|);\,
\bar q_{b,k}\big],
\end{aligned}
\label{eq:judge_input}
\end{equation}
where $\mathbf x_{b,k}\in\mathbb R^{3C+1}$ and
$\operatorname{GAP}$ denotes global average pooling. Let $\mathcal J$ denote the temporal candidate scorer, implemented as a
two-layer MLP with dimensions
$(3C+1)\!\rightarrow\!128\!\rightarrow\!1$. It produces the candidate
logit $a_{b,k}$ as
\begin{equation}
a_{b,k}=\mathcal J(\mathbf x_{b,k}).
\label{eq:judge_logit}
\end{equation}
The current clean teacher BEV constructs the soft target, whereas
$\mathcal J$ receives the scorer input $\mathbf x_{b,k}$. The target and predicted candidate distributions are
\begin{equation}
\begin{aligned}
\omega_{b,k}^{\star}
&= \frac{\exp(z_{b,k}^{\gamma}/\tau)}
{\sum_{m\in\mathcal K_b}\exp(z_{b,m}^{\gamma}/\tau)},
\quad \text{with}\\
z_{b,k}
&= \operatorname{clip}\!\left(
\operatorname{cos}\!\left(
\operatorname{vec}\!\left(
\mathcal W_{b,k}(\mathbf B_{b,k}^{\mathrm{s}})
\right),
\operatorname{vec}(\mathbf B_b^{\mathrm{t}})
\right),0,1\right),\\
\omega_{b,k}
&= \frac{\exp(a_{b,k})}
{\sum_{m\in\mathcal K_b}\exp(a_{b,m})},
\qquad k\in\mathcal K_b,
\end{aligned}
\label{eq:judge_distribution}
\end{equation}
where $z_{b,k}$ denotes the clipped cosine-similarity score between the
aligned historical student BEV and the current clean teacher BEV,
$\operatorname{vec}(\cdot)$ flattens a BEV map, $\gamma>0$ controls target
sharpening, and $\tau>0$ is the temperature parameter. We use
$\gamma=3$ and $\tau=0.1$ in all experiments. Let $\mathcal B_{\mathrm{valid}}$ contain samples with at least one valid
historical candidate. The auxiliary loss is
\begin{equation}
\mathcal L_{\mathrm{judge}}
= \frac{1}{|\mathcal B_{\mathrm{valid}}|}
\sum_{b\in\mathcal B_{\mathrm{valid}}}
\operatorname{KL}\!\left(
\boldsymbol{\omega}_b^{\star}\parallel
\boldsymbol{\omega}_b\right),
\label{eq:judge_loss}
\end{equation}
where $\operatorname{KL}(\cdot\parallel\cdot)$ is the Kullback--Leibler
divergence. The vectors $\boldsymbol{\omega}_b^{\star}$ and
$\boldsymbol{\omega}_b$ collect the corresponding candidate distributions.
If no sample contains a valid historical candidate,
$\mathcal L_{\mathrm{judge}}$ is set to zero.
This loss encourages temporal consistency without changing the reference-point
rectification path or adding a selection module at inference.

\subsection{Training Objective}

NCGR is trained end-to-end with
\begin{equation}
    \mathcal{L}_{\mathrm{total}}
    = \mathcal{L}_{\mathrm{det}}
    + \lambda_{\mathrm{v}}\mathcal{L}_{\mathrm{vanilla}}
    + \lambda_{\mathrm{h}}\mathcal{L}_{\mathrm{health}}
    + \lambda_{\mathrm{j}}\mathcal{L}_{\mathrm{judge}},
    \label{eq:total_loss}
\end{equation}
where $\mathcal L_{\mathrm{det}}$ is the original BEVFormer detection loss,
and the coefficients $\lambda_{\mathrm{v}}$, $\lambda_{\mathrm{h}}$, and
$\lambda_{\mathrm{j}}$ weight the Vanilla BEV distillation,
auxiliary-supervision, and temporal-consistency terms, respectively. The term
$\mathcal L_{\mathrm{vanilla}}$ aligns the perturbed student BEV with the clean
teacher BEV and suppresses offsets on unperturbed cameras;
$\mathcal L_{\mathrm{health}}$ supervises the camera-level control head
and the training-only BEV-quality head; and $\mathcal L_{\mathrm{judge}}$
encourages consistency among
historical BEV features.

\section{Experiments}

\subsection{Experimental Setup}

During NCGR training, synthetic extrinsic perturbations are applied with a
probability of 0.7 to a randomly selected subset of 1--6 cameras. For each selected camera, the Euler-angle and translation components
defining $\Delta\mathbf T_i$ in Eq.~\eqref{eq:compact_perturbation} are sampled
independently from bounded uniform distributions.

Robustness is evaluated under clean, dynamic, and static settings. Clean
evaluation uses the original extrinsics. In dynamic evaluation, a fixed subset
of cameras is used for the entire validation set, and synthetic rigid
perturbations are applied to their extrinsic transformations. For each validation sample, the Euler-angle and translation components
are independently resampled from the same bounded uniform distributions.
All methods use the same nuScenes validation split~\cite{nuscenes} and
perturbations generated with the same random seed. None of the methods has
access to perturbation metadata. Static evaluation
is reported in the supplementary material.

BEVFormer serves as the native SCA baseline, while CAPE is selected as a representative method that reduces reliance on precise extrinsic calibration through camera-local position embeddings. All methods follow their respective standard configurations. NCGR uses BEVFormer-base with a ResNet-101 backbone~\cite{He_2016_CVPR} enhanced by DCNv2~\cite{Zhu_2019_CVPR}. A feature pyramid network (FPN)~\cite{Lin_2017_CVPR} produces the multi-scale image features. The model is trained for 24 epochs. We use $s_{\delta}=0.1$ in all experiments. An inference-time sensitivity analysis over $s_{\delta}\in\{0.05,0.10,0.15,0.20\}$ is reported in the supplementary material. Further implementation details are provided in the supplementary material.

We report the nuScenes detection score (NDS), mean average precision (mAP),
mean average translation error (mATE), mean average scale error (mASE), mean!
average orientation error (mAOE), mean average velocity error (mAVE), and mean
average attribute error (mAAE)~\cite{nuscenes}.

\noindent\textbf{Efficiency.} Efficiency is measured for BEVFormer and NCGR
on an NVIDIA A800 under identical settings with a batch size of 1. NCGR has 69.740M parameters, $1.02\%$ more than the official BEVFormer's
69.035M. Mean latency rises from 205.40 to 217.14 ms/frame ($+5.72\%$).

\subsection{Main Results}

Table~\ref{tab:main_dynamic_15deg} reports results for clean extrinsics and
severe dynamic perturbations with per-axis rotation and translation bounds of
$15^{\circ}$ and $0.1\,\mathrm{m}$, respectively. CAPE~\cite{cape}
achieves the highest NDS and mAP under clean extrinsics. Under perturbations,
NCGR achieves the highest NDS and mAP for each listed fixed camera subset. The
largest margins occur when 3--5 cameras are perturbed. NCGR also achieves the
lowest mAVE for every perturbed subset. When three or more cameras are perturbed,
NCGR achieves its largest margins over the baselines at higher perturbation
severities. Additional quantitative results and qualitative detection visualizations are
provided in the supplementary material.

\begin{table*}[t]
\centering
\begingroup
\small
\setlength{\tabcolsep}{0.5pt}
\begin{tabular}[t]{@{}c l c c c c c c c@{}}
\toprule
\#Cam & Method & NDS$\uparrow$ & mAP$\uparrow$ & mATE$\downarrow$ & mASE$\downarrow$ & mAOE$\downarrow$ & mAVE$\downarrow$ & mAAE$\downarrow$ \\
\midrule
\multirow{3}{*}{Clean}
& BEVF. & 0.518 & 0.417 & 0.673 & 0.273 & 0.371 & 0.393 & 0.198 \\
& CAPE  & \textbf{0.543} & \textbf{0.447} & \textbf{0.643} & \textbf{0.263} & 0.387 & \textbf{0.327} & \textbf{0.185} \\
& NCGR  & 0.520 & 0.415 & 0.697 & 0.270 & \textbf{0.366} & 0.346 & 0.199 \\
\midrule
\multirow{3}{*}{1}
& BEVF. & 0.462 & 0.314 & \textbf{0.678} & 0.279 & \textbf{0.379} & 0.424 & 0.194 \\
& CAPE  & 0.464 & 0.349 & 0.693 & \textbf{0.266} & 0.403 & 0.551 & \textbf{0.193} \\
& NCGR  & \textbf{0.482} & \textbf{0.353} & 0.735 & 0.272 & 0.381 & \textbf{0.357} & 0.201 \\
\midrule
\multirow{3}{*}{2}
& BEVF. & 0.437 & 0.298 & \textbf{0.723} & 0.283 & 0.434 & 0.473 & 0.209 \\
& CAPE  & 0.466 & 0.346 & 0.750 & \textbf{0.267} & \textbf{0.419} & 0.432 & \textbf{0.198} \\
& NCGR  & \textbf{0.479} & \textbf{0.361} & 0.747 & 0.276 & 0.419 & \textbf{0.368} & 0.203 \\
\bottomrule
\end{tabular}\hfill
\begin{tabular}[t]{@{}c l c c c c c c c@{}}
\toprule
\#Cam & Method & NDS$\uparrow$ & mAP$\uparrow$ & mATE$\downarrow$ & mASE$\downarrow$ & mAOE$\downarrow$ & mAVE$\downarrow$ & mAAE$\downarrow$ \\
\midrule
\multirow{3}{*}{3}
& BEVF. & 0.362 & 0.195 & 0.821 & 0.290 & 0.460 & 0.577 & 0.206 \\
& CAPE  & 0.391 & 0.259 & 0.847 & \textbf{0.268} & 0.441 & 0.632 & 0.201 \\
& NCGR  & \textbf{0.451} & \textbf{0.320} & \textbf{0.796} & 0.278 & \textbf{0.426} & \textbf{0.398} & \textbf{0.194} \\
\midrule
\multirow{3}{*}{4}
& BEVF. & 0.355 & 0.197 & \textbf{0.844} & 0.294 & 0.508 & 0.558 & 0.236 \\
& CAPE  & 0.405 & 0.272 & 0.865 & \textbf{0.272} & \textbf{0.444} & 0.512 & 0.221 \\
& NCGR  & \textbf{0.437} & \textbf{0.308} & 0.849 & 0.282 & 0.445 & \textbf{0.388} & \textbf{0.209} \\
\midrule
\multirow{3}{*}{5}
& BEVF. & 0.280 & 0.113 & 0.980 & 0.301 & 0.566 & 0.667 & 0.253 \\
& CAPE  & 0.332 & 0.202 & 0.978 & \textbf{0.272} & \textbf{0.478} & 0.730 & 0.227 \\
& NCGR  & \textbf{0.397} & \textbf{0.256} & \textbf{0.920} & 0.285 & 0.488 & \textbf{0.423} & \textbf{0.195} \\
\bottomrule
\end{tabular}
\par
\parbox{\textwidth}{\centering Each \#Cam row corresponds to one fixed camera subset rather than an average over all subsets of that size; the exact index-to-camera mapping is provided in the supplementary material.}
\endgroup
\caption{Main comparison under clean conditions and severe dynamic perturbations.}
\label{tab:main_dynamic_15deg}
\end{table*}

\begin{figure*}[t]
    \centering
    \includegraphics[width=\textwidth]{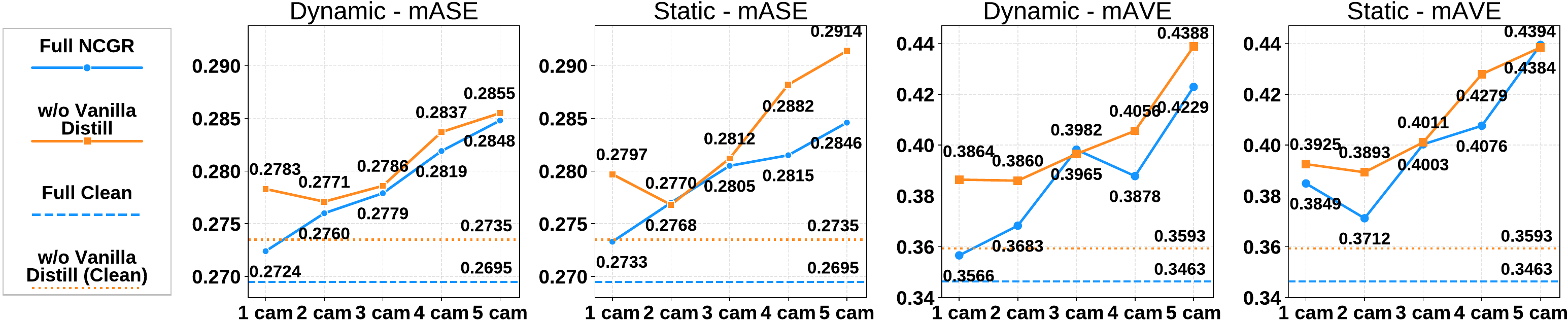}
    \caption{Teacher--student BEV-consistency ablation. The full model lowers
    mASE in all settings and mAVE for clean extrinsics and eight of ten perturbed
    settings, indicating reduced object-scale and velocity errors.}
    \label{fig:vanilla_ablation}
\end{figure*}

\subsection{Ablation Studies}

We examine the two core mechanisms of NCGR through a controlled
inference-time intervention on reference-point rectification and an ablation
of teacher--student BEV consistency.

\paragraph{Controlled intervention on reference-point rectification.}
To isolate the effect of reference-point rectification, we set the learned
effective rectification offset
$\boldsymbol{\delta}_{j,i}^{\star}=g_i\boldsymbol{\delta}_{j,i}$ in
Eq.~\eqref{eq:eff_offset} to zero at inference while keeping all model
parameters fixed. The same validation samples and perturbation realizations
are used, while the native deformable-attention module and all other
computations remain unchanged. As shown in
Table~\ref{tab:ref_intervention}, disabling the effective offset reduces NDS
from 0.3969 to 0.3434 and mAP from 0.2559 to 0.1892, while increasing mAVE
from 0.4229 to 0.4834 under five-camera dynamic perturbations. This performance
drop demonstrates that reference-point rectification contributes substantially
under severe five-camera dynamic perturbations.

\begin{table}[t]
\centering
\begingroup
\small
\setlength{\tabcolsep}{3pt}
\begin{tabular}{lccc}
\toprule
Method & NDS$\uparrow$ & mAP$\uparrow$ & mAVE$\downarrow$ \\
\midrule
NCGR & 0.3969 & 0.2559 & 0.4229 \\
NCGR, offset disabled & 0.3434 & 0.1892 & 0.4834 \\
\bottomrule
\end{tabular}
\endgroup
\caption{Controlled inference-time intervention on reference-point
rectification under the same five-camera dynamic perturbation setting.}
\label{tab:ref_intervention}
\end{table}

\paragraph{Independent gate-endpoint check.}
At inference ($\alpha=1$), we also set $\widehat q_i=1$, yielding
$\rho_i=g_i=0$ and $\boldsymbol{\delta}_{j,i}^{\star}=\mathbf 0$.
A seed-0 evaluation on all 6,019 samples under the same five-camera
dynamic setting yielded 0.3445 NDS, 0.1891 mAP, and 0.5042 mAVE. This independent test confirms that the gate-closed endpoint disables the
effective rectification path, consistent with the direct offset intervention.

\paragraph{Effect of teacher--student BEV consistency.}
We remove the core teacher--student BEV-consistency objective
$\mathcal L_{\mathrm{vanilla}}$ and train the resulting variant using the
same training configuration.
This variant removes foreground-weighted BEV alignment, clean-sample
consistency, and identity regularization, while all other model components and
training settings remain unchanged. We report mASE and mAVE to quantify object-scale and velocity errors,
respectively. As shown in Fig.~\ref{fig:vanilla_ablation}, the full model
achieves lower mASE than the model trained without teacher--student BEV
consistency across all clean, dynamic, and static settings. It also achieves
lower mAVE under clean extrinsics and across eight of the ten perturbed settings.

\section{Conclusion}

In this work, NCGR is proposed to improve the robustness of BEVFormer's
spatial cross-attention to camera extrinsic perturbations, which displace the
image-plane projections of SCA reference points. NCGR addresses these
projection shifts through gated reference-point rectification and BEV
consistency between the clean teacher and perturbed student branches. At
inference, the perturbation-derived training controls are replaced by a learned
camera-level control signal provided by the auxiliary scalar, without requiring
perturbation metadata. Experiments on nuScenes under dynamic and static extrinsic perturbations demonstrate improved robustness over BEVFormer and CAPE, with comparable performance maintained under clean extrinsics.

\bibliography{ncgr_aaai_2027_references}

\onecolumn
\raggedbottom
\setcounter{secnumdepth}{2}
\appendix
\numberwithin{equation}{section}
\numberwithin{figure}{section}
\numberwithin{table}{section}

\begin{center}
{\Large\bfseries Supplementary Material}
\end{center}

This supplementary material is organized around four complementary forms of
evidence. Section~A specifies the NCGR execution path and the complete
extrinsic-perturbation protocol. Section~B documents the implementation,
evaluation, efficiency-benchmarking, and sensitivity-analysis settings required
to reproduce the reported results. Section~C provides the complete robustness tables and a
class-wise breakdown under the strongest matched perturbation setting.
Section~D isolates the rectification path through controlled interventions,
provides qualitative evidence at the detection and sampling levels, and
quantifies whether the camera- and layer-conditioned corrections reduce
projection error relative to clean references.

\section{Additional Method and Control Details}
\label{app:protocol_details}

\subsection{NCGR Execution Path inside Spatial Cross-Attention}
NCGR changes the geometric anchor used by spatial cross-attention (SCA) while
preserving the native multi-scale deformable sampling operation. For BEV query
$j$, camera $i$, and height anchor $d$, we follow the notation defined in the
main paper: $\mathbf r_{j,i,d}$ is the normalized image-plane reference point,
$\mathbf c_i$ is the camera-level condition, and $g_i$ is the rectification
gate. The $i$-th 2D output slice of the shared correction network predicts the
bounded offset, and NCGR rectifies the reference point as
\begin{equation}
\begin{aligned}
\boldsymbol{\delta}_{j,i}
&=s_{\delta}\tanh\!\left(
\mathcal G^{(i)}([\mathbf Q_j;\mathbf c_i])
\right),\\
\boldsymbol{\delta}_{j,i}^{\star}
&=g_i\boldsymbol{\delta}_{j,i},\\
\mathbf r_{j,i,d}^{\mathrm{rect}}
&=\mathbf r_{j,i,d}+\boldsymbol{\delta}_{j,i}^{\star}.
\end{aligned}
\label{eq:app_ncgr_execution}
\end{equation}
The scheduled construction of $\mathbf c_i$ and $g_i$ is unchanged from the
main paper. At inference, $\alpha=1$, so that
$\mathbf c_i=\rho_i\mathbf 1_2$, $g_i=\rho_i$, and
$\rho_i=1-\operatorname{sg}(\widehat q_i)$. All quantities in
Eq.~\eqref{eq:app_ncgr_execution} are expressed in the normalized image-plane
coordinates used by SCA. The same effective offset
$\boldsymbol{\delta}_{j,i}^{\star}$ is applied to all height anchors associated
with the same query--camera pair. The rectified reference point is then passed
to the unchanged deformable-attention sampler, which adds its native head-,
level-, and point-specific offsets around that point. Thus, NCGR first
compensates for the projection-anchor displacement and then retains the local
feature search of the original SCA implementation.

\begin{table}[H]
\centering
\begingroup
\small
\setlength{\tabcolsep}{3.5pt}
\begin{tabular}{@{}p{0.34\linewidth}p{0.56\linewidth}@{}}
\toprule
\textbf{Stage} & \textbf{Operation} \\
\midrule
Base projection & Project each BEV reference point using the extrinsics
available to the detector. \\
Camera control & Construct $\mathbf c_i$ and $g_i$ using the scheduled
camera-level controls; at inference both are determined by $\widehat q_i$. \\
Anchor rectification & Predict and gate one bounded 2D residual for each
query--camera pair. \\
Native sampling & Apply the original multi-scale deformable offsets around the
rectified anchor. \\
\bottomrule
\end{tabular}
\endgroup
\caption{Execution order of NCGR inside SCA. The first three stages determine
the rectified geometric anchor; native deformable sampling is retained as the
final stage.}
\label{tab:app_execution_path}
\end{table}

\subsection{Extrinsic Perturbation Protocol}
This section provides the complete projection and perturbation formulation
summarized in the main paper. For a 3D reference point
$\mathbf p=[p_x,p_y,p_z]^\top$, the calibrated projection into the image plane
of camera $i$ is
\begin{equation}
\begin{aligned}
\mathbf u_i
&=
\mathbf T_{\mathrm{l2i}}^{(i)}
\begin{bmatrix}\mathbf p\\1\end{bmatrix},\\
\boldsymbol{\pi}_i(\mathbf p)
&=
\left(
\frac{u_{i,x}}{u_{i,z}},
\frac{u_{i,y}}{u_{i,z}}
\right)\in\mathbb R^2,\\
\mathbf T_{\mathrm{l2i}}^{(i)}
&=
\mathbf K_i[\,\mathbf R_i\mid\mathbf t_i\,],
\end{aligned}
\label{eq:app_projection}
\end{equation}
where
$\mathbf T_{\mathrm{l2i}}^{(i)}\in\mathbb R^{3\times4}$ is the
LiDAR-to-image projection matrix,
$\mathbf u_i=[u_{i,x},u_{i,y},u_{i,z}]^\top\in\mathbb R^3$,
$\mathbf K_i\in\mathbb R^{3\times3}$ is the camera intrinsic matrix, and
$\mathbf R_i\in\mathrm{SO}(3)$ and $\mathbf t_i\in\mathbb R^3$ are the
rotation and translation of the calibrated LiDAR-to-camera transformation.

The homogeneous LiDAR-to-camera transformation and the synthetic rigid
perturbation are
\begin{equation}
\mathbf T_{\mathrm{l2c}}^{(i)}
=
\begin{bmatrix}
\mathbf R_i & \mathbf t_i\\
\mathbf 0^\top & 1
\end{bmatrix},
\qquad
\Delta\mathbf T_i
=
\begin{bmatrix}
\Delta\mathbf R_i & \Delta\mathbf t_i\\
\mathbf 0^\top & 1
\end{bmatrix}
\in\mathrm{SE}(3).
\label{eq:app_transform_definitions}
\end{equation}
Here, $\Delta\mathbf R_i\in\mathrm{SO}(3)$ and
$\Delta\mathbf t_i\in\mathbb R^3$ denote the rotational and translational
components of the synthetic perturbation, respectively.
A perturbation is applied by left multiplication:
\begin{equation}
\widetilde{\mathbf T}_{\mathrm{l2c}}^{(i)}
=
\Delta\mathbf T_i\mathbf T_{\mathrm{l2c}}^{(i)}.
\label{eq:app_perturb}
\end{equation}
Under the column-vector convention, the perturbed rotation and translation are
\[
\widetilde{\mathbf R}_i
=
\Delta\mathbf R_i\mathbf R_i,
\qquad
\widetilde{\mathbf t}_i
=
\Delta\mathbf R_i\mathbf t_i+\Delta\mathbf t_i.
\]
The rotational perturbation uses the roll, pitch, and yaw parameterization
\[
\boldsymbol{\alpha}_i=[\phi_i,\vartheta_i,\psi_i]^\top,
\qquad
\Delta\mathbf R_i
=
\mathbf R_z(\psi_i)\mathbf R_y(\vartheta_i)\mathbf R_x(\phi_i),
\]
where $\phi_i$, $\vartheta_i$, and $\psi_i$ denote the roll, pitch, and yaw
angles, respectively, and $\mathbf R_x$, $\mathbf R_y$, and $\mathbf R_z$
denote rotations about the corresponding camera axes. We use the standard
camera-axis convention, in which the $x$-, $y$-, and $z$-axes point right,
down, and forward, respectively. The axis-rotation matrices are
\[
\begin{aligned}
\mathbf R_x(\phi)
&=
\begin{bmatrix}
1&0&0\\
0&\cos\phi&-\sin\phi\\
0&\sin\phi&\cos\phi
\end{bmatrix},\\[4pt]
\mathbf R_y(\vartheta)
&=
\begin{bmatrix}
\cos\vartheta&0&\sin\vartheta\\
0&1&0\\
-\sin\vartheta&0&\cos\vartheta
\end{bmatrix},\\[4pt]
\mathbf R_z(\psi)
&=
\begin{bmatrix}
\cos\psi&-\sin\psi&0\\
\sin\psi&\cos\psi&0\\
0&0&1
\end{bmatrix}.
\end{aligned}
\]

Combining
$\widetilde{\mathbf T}_{\mathrm{l2c}}^{(i)}$ with $\mathbf K_i$ gives the
perturbed LiDAR-to-image projection and the image-plane location
$\widetilde{\boldsymbol{\pi}}_i(\mathbf p)$. The resulting displacement is
\begin{equation}
\Delta\boldsymbol{\pi}_i(\mathbf p)
=
\widetilde{\boldsymbol{\pi}}_i(\mathbf p)
-
\boldsymbol{\pi}_i(\mathbf p),
\qquad
\Delta\boldsymbol{\pi}_i(\mathbf p)\in\mathbb R^2.
\label{eq:app_projection_displacement}
\end{equation}
This displacement moves the SCA base projection to a perturbed image-plane
location. The reference-point rectification branch compensates for this projection
error before native deformable sampling.

\begin{figure}[H]
    \centering
    \includegraphics[width=0.90\linewidth]{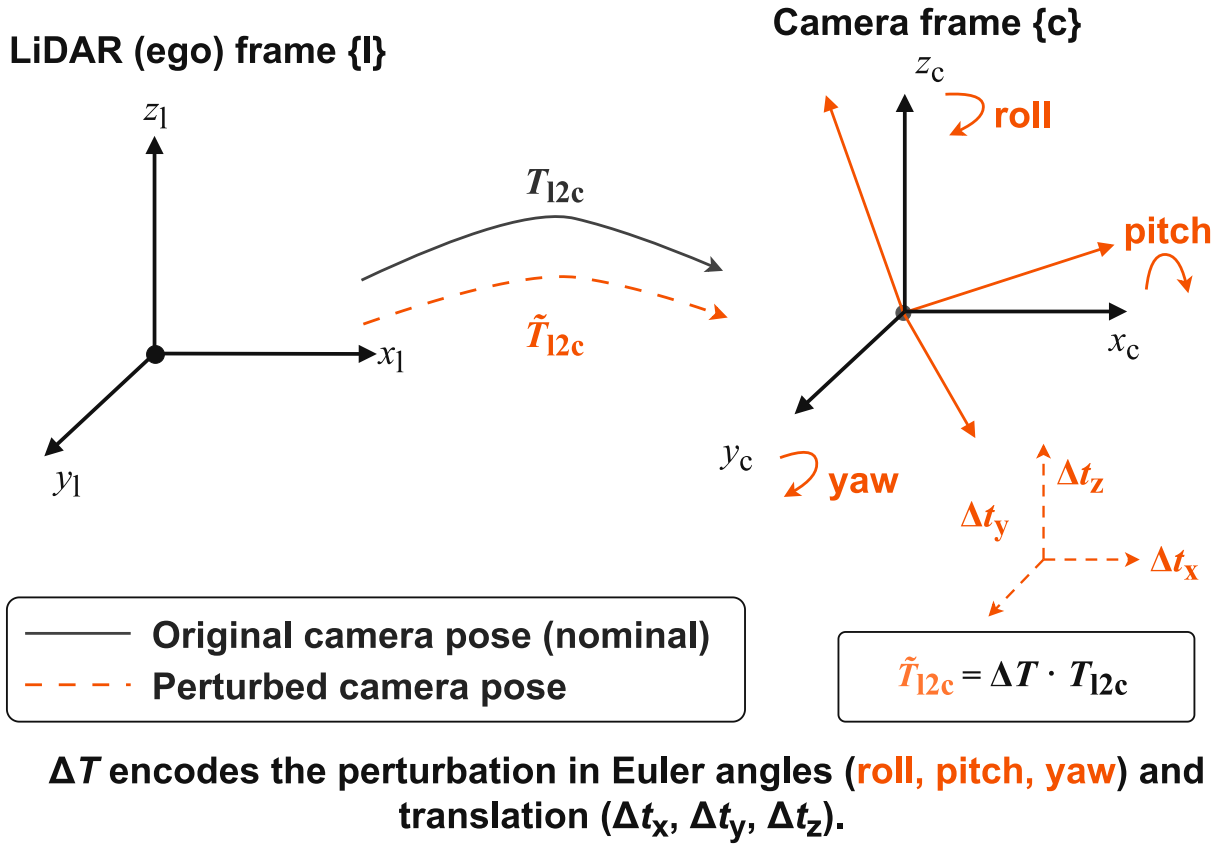}
    \caption{Extrinsic-perturbation protocol. A left-multiplied rigid
    perturbation is applied to the LiDAR-to-camera transformation, and the
    resulting extrinsics are combined with the camera intrinsics to obtain
    the perturbed LiDAR-to-image projection used by SCA.}
    \label{fig:app_extrinsic_protocol}
\end{figure}

During NCGR training, the roll, pitch, and yaw angles are sampled independently from
$\mathcal U(-15^\circ,15^\circ)$, while the three translation components are
sampled independently from $\mathcal U(-0.1,0.1)\,\mathrm m$, where
$\mathcal U(a,b)$ denotes the continuous uniform distribution on $[a,b]$.
In Fig.~\ref{fig:app_extrinsic_protocol}, $\{l\}$ and $\{c\}$ denote the LiDAR
(ego) and camera coordinate frames, respectively, and the camera index $i$ is
suppressed; $\Delta t_x$, $\Delta t_y$, and $\Delta t_z$ denote the three
components of $\Delta\mathbf t_i$. During the
severity sweep, the translation bound remains $0.1\,\mathrm m$ and the
rotation bound is set to $3^\circ$, $6^\circ$, $9^\circ$, $12^\circ$, or
$15^\circ$. The synthetic perturbations are used to construct
clean--perturbed training pairs and auxiliary training signals, but are not
provided during inference. Figure~\ref{fig:app_extrinsic_protocol} summarizes
the process. The dynamic and static resampling rules and the information withheld
during blind inference are specified in the next subsection.

\subsection{Blind Evaluation Protocol}
Table~\ref{tab:appendix_protocol_summary} summarizes the evaluation protocol.
Dynamic perturbations are resampled for every validation sample. In the static
setting, each selected camera receives one fixed maximum-magnitude perturbation:
the sign of each rotational and translational component is sampled once and
held fixed throughout validation. The same realization is reused across methods.
Training perturbation metadata is used only to construct the condition and gate
and to supervise the auxiliary camera-level scalar. At inference, the indices of the perturbed cameras and the perturbation
magnitudes are unavailable; the
condition and gate are generated from the learned scalar.

\begin{table}[H]
\centering
\begingroup
\small
\setlength{\tabcolsep}{3.5pt}
\begin{tabular}{@{}p{0.31\linewidth}p{0.61\linewidth}@{}}
\toprule
\textbf{Item} & \textbf{Setting} \\
\midrule
Perturbed geometry & Left-multiplied LiDAR-to-camera perturbation:
$\widetilde{\mathbf T}_{\mathrm{l2c}}^{(i)}
=\Delta\mathbf T_i\mathbf T_{\mathrm{l2c}}^{(i)}$. \\
Rotation & Independent roll, pitch, and yaw perturbations with per-axis bounds
of $3^\circ$, $6^\circ$, $9^\circ$, $12^\circ$, or $15^\circ$. \\
Translation & Independent $x$, $y$, and $z$ perturbations with a per-axis bound
of $0.1$ m. \\
Dynamic setting & Resampled for each validation sample. \\
Static setting & One maximum-magnitude realization is sampled once, shared
across methods, and fixed throughout validation. \\
Camera settings & For each camera-count setting, one predefined camera subset
of the corresponding size is used. \\
Training-time signals & Synthetic perturbations are used to construct the
condition and gate and to supervise the auxiliary camera-level scalar. \\
Inference & The identities of the perturbed cameras and the perturbation
magnitudes are unavailable; blind inference uses the learned scalar. \\
\bottomrule
\end{tabular}
\endgroup
\caption{Summary of the extrinsic perturbation and blind evaluation protocol.}
\label{tab:appendix_protocol_summary}
\end{table}

\subsection{Fixed Camera Subsets}
Table~\ref{tab:camera_subset_mapping} lists the fixed camera subset
used for each camera-count setting throughout the robustness evaluation.

\begin{table}[H]
\centering
\begingroup
\small
\setlength{\tabcolsep}{3.5pt}
\begin{tabular}{@{}lp{0.72\linewidth}@{}}
\toprule
\textbf{Setting} & \textbf{nuScenes cameras} \\
\midrule
1 cam & \texttt{CAM\_FRONT} \\
2 cam & \texttt{CAM\_FRONT\_RIGHT}, \texttt{CAM\_FRONT\_LEFT} \\
3 cam & \texttt{CAM\_FRONT\_RIGHT}, \texttt{CAM\_FRONT\_LEFT}, \texttt{CAM\_BACK} \\
4 cam & \texttt{CAM\_FRONT\_RIGHT}, \texttt{CAM\_FRONT\_LEFT}, \texttt{CAM\_BACK\_LEFT}, \texttt{CAM\_BACK\_RIGHT} \\
5 cam & All cameras except \texttt{CAM\_FRONT} \\
\bottomrule
\end{tabular}
\endgroup
\caption{Fixed camera-count settings used in the robustness tables.}
\label{tab:camera_subset_mapping}
\end{table}

\subsection{Paired Evaluation and Evaluation Seeds}
All methods use the same nuScenes validation split, predefined camera subsets,
and matched perturbations; the static realization is also shared across methods.
The main comparison tables use evaluation seed 0.
Section~\ref{app:seed_sensitivity} evaluates the same NCGR checkpoint using
seed 0 and three additional evaluation-time seeds under the predefined
five-camera stress test with per-axis rotation and translation bounds of
$15^\circ$ and $0.1$ m, respectively.

\FloatBarrier
\section{Implementation and Reproducibility}
\label{app:efficiency_seed}

\subsection{Implementation and Training Configuration}
The comparison methods follow their respective standard configurations. NCGR
uses a ResNet-101 backbone with deformable convolutions and an FPN image
encoder. The BEV representation contains a $200\times200$ grid, 900 
queries, six SCA encoder layers, and four height anchors per BEV reference
point. The rectification scale is $s_{\delta}=0.10$ in the main configuration.

Training uses AdamW for 24 epochs with an initial learning rate of
$2\times10^{-4}$, weight decay $0.01$, and a backbone learning-rate multiplier
of $0.1$. Gradients are clipped to a maximum norm of 35. The cosine learning-rate
schedule uses 500 linear warm-up iterations with a warm-up ratio of $1/3$ and a
minimum learning-rate ratio of $10^{-3}$. The training seed is fixed to 0 and
deterministic execution is enabled. Training starts from the
ResNet-101-DCNv2 FCOS3D initialization used by the corresponding BEVFormer
configuration.

The extrinsic simulator is applied with probability 0.7. When activated, it
selects between one and six cameras and samples independent roll, pitch, and
yaw perturbations from $\mathcal U(-15^\circ,15^\circ)$ and translation
components from $\mathcal U(-0.1,0.1)\,\mathrm m$. The loss weights are
$\lambda_v=0.2$, $\lambda_{\mathrm{clean}}=0.25$,
$\lambda_{\mathrm{id}}=0.5$, $\lambda_h=0.15$,
$\lambda_{\mathrm{cam}}=1.0$, and $\lambda_j=0.1$. Thus, the effective weights
of $\mathcal L_{\mathrm{align}}$, $\mathcal L_{\mathrm{clean}}$, and
$\mathcal L_{\mathrm{id}}$ are 0.2, 0.05, and 0.1, respectively.

\begin{center}
\begin{minipage}{0.96\linewidth}
\centering
\begingroup
\small
\setlength{\tabcolsep}{3.5pt}
\begin{tabular}{@{}p{0.42\linewidth}p{0.48\linewidth}@{}}
\toprule
\textbf{Item} & \textbf{Setting} \\
\midrule
Backbone & ResNet-101-DCNv2 with FPN \\
BEV grid / queries & $200\times200$ / 900 \\
SCA layers / height anchors & 6 / 4 \\
Optimizer & AdamW; learning rate $2\times10^{-4}$; weight decay $0.01$ \\
Schedule & Cosine, 500-iteration linear warm-up \\
Training length & 24 epochs \\
Gradient clipping & Maximum norm 35 \\
Training seed & 0, deterministic execution \\
Perturbation probability & 0.7 \\
Perturbed cameras per sample & Between 1 and 6 when activated \\
Main offset scale & $s_{\delta}=0.10$ \\
\bottomrule
\end{tabular}
\endgroup
\captionof{table}{Principal implementation and training settings for NCGR.}
\label{tab:app_implementation}
\end{minipage}
\end{center}

\subsection{Runtime and Memory Efficiency}
\label{app:efficiency}

Official BEVFormer, a runtime-disabled NCGR control, and the full NCGR model are
benchmarked on an NVIDIA A800 80 GB PCIe GPU with batch size 1. All methods use
32-bit floating-point (FP32) execution with TensorFloat-32 (TF32) enabled
consistently. The input tensor has shape $[1,6,3,928,1600]$, ordered as batch
size, number of cameras, channels, image height, and image width; the detector
uses a $200\times200$ BEV grid and 900 queries.
For each method, 50 warm-up frames are discarded and 500 frames are measured
in each of three independent processes, giving 1,500 measured frames in total.
Reported latency includes CPU-to-GPU scatter, detector forward propagation,
and bounding-box decoding; data loading and image preprocessing are excluded.
P50 and P95 denote the 50th- and 95th-percentile latency, respectively, and
FPS denotes throughput in frames per second.

\begin{table}[H]
\centering
\begingroup
\small
\setlength{\tabcolsep}{2.4pt}
\begin{tabular}{@{}lcccccc@{}}
\toprule
Method &
\shortstack{Parameters\\(million)} &
\shortstack{Mean latency\\(ms)} &
P50 (ms) &
P95 (ms) &
FPS &
\shortstack{Peak allocated\\memory (GiB)} \\
\midrule
Official BEVFormer &
69.035 &
$205.399 \pm 1.024$ &
$205.110 \pm 0.913$ &
$207.555 \pm 1.585$ &
$4.8686 \pm 0.0242$ &
2.125 \\
NCGR runtime-disabled control &
69.740 &
$204.303 \pm 0.210$ &
$204.127 \pm 0.189$ &
$205.451 \pm 0.497$ &
$4.8947 \pm 0.0050$ &
2.127 \\
Full NCGR &
69.740 &
$217.140 \pm 0.748$ &
$216.969 \pm 0.738$ &
$218.952 \pm 1.015$ &
$4.6053 \pm 0.0159$ &
2.141 \\
\bottomrule
\end{tabular}
\endgroup
\caption{Runtime and memory efficiency on an NVIDIA A800. Latency and FPS are reported as the mean $\pm$ standard deviation across three processes.}
\label{tab:app_efficiency_full}
\end{table}

\begin{table}[H]
\centering
\begingroup
\small
\setlength{\tabcolsep}{3pt}
\begin{tabular}{@{}p{0.27\linewidth}p{0.40\linewidth}p{0.23\linewidth}@{}}
\toprule
Metric & Absolute change & Relative change \\
\midrule
Mean latency & $+11.741$ ms/frame & $+5.72\%$ \\
FPS & $-0.2633$ FPS & $-5.41\%$ \\
Peak allocated memory & $+16.0$ MiB & $+0.74\%$ \\
Parameters & $+0.705$ million & $+1.02\%$ \\
\bottomrule
\end{tabular}
\endgroup
\caption{Full NCGR overhead relative to official BEVFormer.}
\label{tab:app_efficiency_overhead}
\end{table}

The full NCGR model runs at 217.14 ms/frame (4.605 FPS) and uses 2.141 GiB of peak
allocated GPU memory, compared with 205.40 ms/frame (4.869 FPS) and
2.125 GiB for official BEVFormer. The latency difference between the full NCGR model and the runtime-disabled
control is 12.837 ms/frame (6.28\%). This control is used only to isolate the runtime cost of executing the auxiliary camera-level head and
correction network within the same NCGR implementation.

\subsection{Inference-Time Sensitivity to the Offset Scale}
\label{app:sdelta_sensitivity}

To assess sensitivity to the maximum rectification offset, we vary
$s_{\delta}$ only at inference while keeping the trained NCGR checkpoint and
all other evaluation settings fixed. Here, $s_{\delta}$ bounds each component
of the learned offset in the normalized image-plane coordinates used by SCA.
We evaluate $s_{\delta}\in\{0.05,0.10,0.15,0.20\}$ under clean extrinsics and
the predefined five-camera dynamic stress test with per-axis rotation and
translation bounds of $15^\circ$ and $0.1$ m, respectively.

\begin{table}[H]
\centering
\begingroup
\small
\setlength{\tabcolsep}{4.5pt}
\begin{tabular}{@{}ccccc@{}}
\toprule
& \multicolumn{2}{c}{Clean} & \multicolumn{2}{c}{Dynamic, 5 cam} \\
\cmidrule(lr){2-3}\cmidrule(l){4-5}
$s_{\delta}$ & NDS$\uparrow$ & mAP$\uparrow$ & NDS$\uparrow$ & mAP$\uparrow$ \\
\midrule
0.05 & 0.4996 & 0.3908 & 0.3784 & 0.2331 \\
0.10 & \textbf{0.5199} & \textbf{0.4154} & \textbf{0.3969} & \textbf{0.2559} \\
0.15 & 0.4817 & 0.3604 & 0.3828 & 0.2436 \\
0.20 & 0.3658 & 0.1982 & 0.3352 & 0.1708 \\
\bottomrule
\end{tabular}
\endgroup
\caption{Inference-time sensitivity of NCGR to the rectification-offset scale
$s_{\delta}$. The same trained checkpoint is used in all rows, and the best
result in each column is shown in bold.}
\label{tab:app_sdelta_sensitivity}
\end{table}

Within the evaluated range, $s_{\delta}=0.10$ achieves the highest NDS and
mAP under both clean and perturbed conditions. Smaller or larger values yield
lower performance, with a pronounced degradation at $s_{\delta}=0.20$. These results
support $s_{\delta}=0.10$ as the operating point used in the main experiments.

\subsection{Evaluation-Seed Sensitivity}
\label{app:seed_sensitivity}

The main comparison tables use evaluation seed 0. To assess sensitivity to
evaluation-time perturbation realizations, the same epoch-24 NCGR checkpoint
is evaluated using seed 0 and three additional seeds: 3407, 104729, and
20260715. For every run, both the evaluation seed and
\texttt{PYTHONHASHSEED} are fixed. The experiment uses per-axis rotation and translation bounds of $15^\circ$
and $0.1$ m, respectively, in the predefined five-camera setting. Dynamic and static perturbations follow the protocol
in Section~\ref{app:protocol_details}.

\begin{table}[H]
\centering
\begingroup
\small
\setlength{\tabcolsep}{5pt}
\begin{tabular}{@{}rcc@{}}
\toprule
Seed & Dynamic NDS & Dynamic mAP \\
\midrule
0 (main tables) & 0.39690 & 0.25590 \\
3407            & 0.39725 & 0.25360 \\
104729          & 0.39912 & 0.25887 \\
20260715        & 0.39674 & 0.25549 \\
\midrule
Mean $\pm$ sample std. &
$0.39750 \pm 0.00110$ &
$0.25597 \pm 0.00218$ \\
\bottomrule
\end{tabular}

\medskip
\begin{tabular}{@{}rcc@{}}
\toprule
Seed & Static NDS & Static mAP \\
\midrule
0 (main tables) & 0.33730 & 0.15870 \\
3407            & 0.33639 & 0.15751 \\
104729          & 0.33156 & 0.15886 \\
20260715        & 0.33038 & 0.15938 \\
\midrule
Mean $\pm$ sample std. &
$0.33391 \pm 0.00345$ &
$0.15861 \pm 0.00079$ \\
\bottomrule
\end{tabular}
\endgroup
\caption{Evaluation-seed sensitivity of NCGR under the predefined
five-camera $15^\circ$/0.1 m stress test. Seed 0 is used in the main
comparison tables. The final row reports the mean and sample standard deviation
over all four evaluation seeds.}
\label{tab:app_seed_sensitivity}
\end{table}

Across the four seeds, the ranges of Dynamic NDS, Dynamic mAP, Static NDS,
and Static mAP are 0.00238, 0.00527, 0.00692, and 0.00187, respectively.
The small variation among the four runs indicates stable performance over
the tested perturbation realizations in the predefined five-camera setting with
per-axis rotation and translation bounds of $15^\circ$ and $0.1$ m, respectively.

\FloatBarrier
\section{Complete Robustness Results}
\label{app:full_metrics}

Additional robustness results are reported below. The translation bound is
fixed at 0.1 m. Dynamic results cover per-axis rotation bounds from
$3^\circ$ to $12^\circ$, while static results additionally include
$15^\circ$. The clean reference and the $15^\circ$ dynamic setting are
already reported in the main paper and are therefore omitted here. Dynamic
and static results use the fixed camera subsets in
Table~\ref{tab:camera_subset_mapping}.

\subsection{Summary of Robustness Trends}
The full results reveal severity-dependent trends. NCGR improves NDS
and mAP over official BEVFormer across all evaluated fixed camera subsets
and perturbation levels. NCGR's advantage over CAPE becomes more pronounced as
the perturbation severity and the number of affected cameras increase,
particularly at rotation bounds of $9^\circ$--$15^\circ$ when at least
three cameras are perturbed. NCGR also achieves substantially lower mAVE in
many severe perturbation settings. Its gains
are most pronounced under moderate-to-severe perturbations.

\begin{table}[H]
\centering
\begingroup
\small
\renewcommand{\arraystretch}{0.65}
\setlength{\tabcolsep}{10pt}
\begin{tabular}{@{}cllccccccc@{}}
\toprule
Bound & \#Cam & Method & NDS$\uparrow$ & mAP$\uparrow$ & mATE$\downarrow$ & mASE$\downarrow$ & mAOE$\downarrow$ & mAVE$\downarrow$ & mAAE$\downarrow$ \\
\midrule
\multirow{15}{*}{$3^\circ$/0.1 m} & \multirow{3}{*}{1 cam} & BEVFormer & 0.4994 & 0.3953 & 0.7072 & 0.2746 & 0.3814 & 0.4228 & 0.1966 \\
&  & CAPE & \textbf{0.5250} & \textbf{0.4292} & \textbf{0.6835} & \textbf{0.2636} & 0.3894 & 0.3740 & \textbf{0.1857} \\
&  & \textbf{NCGR} & 0.5062 & 0.3960 & 0.7305 & 0.2701 & \textbf{0.3671} & \textbf{0.3515} & 0.1995 \\
\cmidrule(l){2-10}
& \multirow{3}{*}{2 cam} & BEVFormer & 0.5018 & 0.3975 & 0.7037 & 0.2745 & \textbf{0.3714} & 0.4209 & 0.1996 \\
&  & CAPE & \textbf{0.5252} & \textbf{0.4276} & \textbf{0.6893} & \textbf{0.2639} & 0.3941 & \textbf{0.3522} & \textbf{0.1871} \\
&  & \textbf{NCGR} & 0.5091 & 0.4017 & 0.7175 & 0.2724 & 0.3749 & 0.3527 & 0.2001 \\
\cmidrule(l){2-10}
& \multirow{3}{*}{3 cam} & BEVFormer & 0.4876 & 0.3837 & 0.7378 & 0.2752 & \textbf{0.3746} & 0.4510 & 0.2033 \\
&  & CAPE & \textbf{0.5097} & \textbf{0.4120} & \textbf{0.7238} & \textbf{0.2637} & 0.3938 & 0.3880 & \textbf{0.1936} \\
&  & \textbf{NCGR} & 0.5001 & 0.3912 & 0.7405 & 0.2721 & 0.3792 & \textbf{0.3650} & 0.1986 \\
\cmidrule(l){2-10}
& \multirow{3}{*}{4 cam} & BEVFormer & 0.4882 & 0.3820 & 0.7404 & 0.2759 & \textbf{0.3783} & 0.4305 & 0.2031 \\
&  & CAPE & \textbf{0.5122} & \textbf{0.4120} & \textbf{0.7250} & \textbf{0.2634} & 0.3962 & 0.3617 & \textbf{0.1914} \\
&  & \textbf{NCGR} & 0.4996 & 0.3895 & 0.7411 & 0.2736 & 0.3817 & \textbf{0.3542} & 0.2015 \\
\cmidrule(l){2-10}
& \multirow{3}{*}{5 cam} & BEVFormer & 0.4715 & 0.3675 & 0.7835 & 0.2769 & 0.3931 & 0.4648 & 0.2048 \\
&  & CAPE & \textbf{0.4970} & \textbf{0.3976} & \textbf{0.7632} & \textbf{0.2641} & 0.3982 & 0.3962 & \textbf{0.1965} \\
&  & \textbf{NCGR} & 0.4896 & 0.3772 & 0.7710 & 0.2737 & \textbf{0.3770} & \textbf{0.3676} & 0.2006 \\
\midrule
\multirow{15}{*}{$6^\circ$/0.1 m} & \multirow{3}{*}{1 cam} & BEVFormer & 0.4797 & 0.3642 & 0.7144 & 0.2767 & 0.3811 & 0.4507 & 0.2013 \\
&  & CAPE & \textbf{0.5026} & \textbf{0.4022} & \textbf{0.7027} & \textbf{0.2638} & 0.3939 & 0.4352 & \textbf{0.1893} \\
&  & \textbf{NCGR} & 0.4914 & 0.3711 & 0.7408 & 0.2719 & \textbf{0.3756} & \textbf{0.3540} & 0.1988 \\
\cmidrule(l){2-10}
& \multirow{3}{*}{2 cam} & BEVFormer & 0.4774 & 0.3645 & 0.7319 & 0.2775 & 0.3995 & 0.4363 & 0.2037 \\
&  & CAPE & \textbf{0.5035} & \textbf{0.4007} & \textbf{0.7294} & \textbf{0.2650} & 0.4032 & 0.3771 & \textbf{0.1934} \\
&  & \textbf{NCGR} & 0.4937 & 0.3812 & 0.7467 & 0.2747 & \textbf{0.3860} & \textbf{0.3572} & 0.2037 \\
\cmidrule(l){2-10}
& \multirow{3}{*}{3 cam} & BEVFormer & 0.4385 & 0.3223 & 0.8100 & 0.2800 & 0.4140 & 0.5116 & 0.2108 \\
&  & CAPE & 0.4664 & \textbf{0.3612} & 0.8012 & \textbf{0.2644} & 0.4018 & 0.4735 & 0.2007 \\
&  & \textbf{NCGR} & \textbf{0.4729} & 0.3528 & \textbf{0.7934} & 0.2755 & \textbf{0.3924} & \textbf{0.3740} & \textbf{0.1995} \\
\cmidrule(l){2-10}
& \multirow{3}{*}{4 cam} & BEVFormer & 0.4414 & 0.3232 & 0.8135 & 0.2805 & 0.4184 & 0.4748 & 0.2147 \\
&  & CAPE & \textbf{0.4749} & \textbf{0.3648} & \textbf{0.7997} & \textbf{0.2651} & \textbf{0.4046} & 0.4029 & \textbf{0.2026} \\
&  & \textbf{NCGR} & 0.4684 & 0.3516 & 0.8040 & 0.2775 & 0.4186 & \textbf{0.3681} & 0.2056 \\
\cmidrule(l){2-10}
& \multirow{3}{*}{5 cam} & BEVFormer & 0.4006 & 0.2800 & 0.9054 & 0.2837 & 0.4380 & 0.5498 & 0.2170 \\
&  & CAPE & 0.4362 & \textbf{0.3293} & 0.8791 & \textbf{0.2653} & \textbf{0.4212} & 0.5065 & 0.2124 \\
&  & \textbf{NCGR} & \textbf{0.4458} & 0.3226 & \textbf{0.8653} & 0.2791 & 0.4333 & \textbf{0.3815} & \textbf{0.1956} \\
\midrule
\multirow{15}{*}{$9^\circ$/0.1 m} & \multirow{3}{*}{1 cam} & BEVFormer & 0.4705 & 0.3399 & \textbf{0.6987} & 0.2772 & \textbf{0.3773} & 0.4431 & 0.1982 \\
&  & CAPE & 0.4851 & \textbf{0.3793} & 0.7039 & \textbf{0.2643} & 0.3990 & 0.4862 & \textbf{0.1919} \\
&  & \textbf{NCGR} & \textbf{0.4855} & 0.3615 & 0.7363 & 0.2723 & 0.3803 & \textbf{0.3612} & 0.2020 \\
\cmidrule(l){2-10}
& \multirow{3}{*}{2 cam} & BEVFormer & 0.4583 & 0.3363 & \textbf{0.7339} & 0.2808 & 0.4169 & 0.4578 & 0.2093 \\
&  & CAPE & \textbf{0.4876} & \textbf{0.3777} & 0.7428 & \textbf{0.2659} & 0.4096 & 0.3973 & \textbf{0.1975} \\
&  & \textbf{NCGR} & 0.4866 & 0.3703 & 0.7551 & 0.2751 & \textbf{0.3890} & \textbf{0.3628} & 0.2033 \\
\cmidrule(l){2-10}
& \multirow{3}{*}{3 cam} & BEVFormer & 0.4037 & 0.2671 & 0.8396 & 0.2833 & 0.4371 & 0.5282 & 0.2103 \\
&  & CAPE & 0.4331 & 0.3186 & 0.8324 & \textbf{0.2653} & 0.4160 & 0.5451 & 0.2025 \\
&  & \textbf{NCGR} & \textbf{0.4633} & \textbf{0.3366} & \textbf{0.7930} & 0.2766 & \textbf{0.4080} & \textbf{0.3813} & \textbf{0.1912} \\
\cmidrule(l){2-10}
& \multirow{3}{*}{4 cam} & BEVFormer & 0.4010 & 0.2683 & 0.8389 & 0.2868 & 0.4592 & 0.5214 & 0.2255 \\
&  & CAPE & 0.4449 & 0.3264 & 0.8354 & \textbf{0.2678} & \textbf{0.4166} & 0.4484 & 0.2150 \\
&  & \textbf{NCGR} & \textbf{0.4539} & \textbf{0.3296} & \textbf{0.8299} & 0.2796 & 0.4227 & \textbf{0.3739} & \textbf{0.2030} \\
\cmidrule(l){2-10}
& \multirow{3}{*}{5 cam} & BEVFormer & 0.3412 & 0.2047 & 0.9703 & 0.2910 & 0.4988 & 0.6216 & 0.2297 \\
&  & CAPE & 0.3912 & 0.2757 & 0.9356 & \textbf{0.2672} & \textbf{0.4433} & 0.5979 & 0.2220 \\
&  & \textbf{NCGR} & \textbf{0.4236} & \textbf{0.2899} & \textbf{0.8806} & 0.2807 & 0.4521 & \textbf{0.4032} & \textbf{0.1969} \\
\midrule
\multirow{15}{*}{$12^\circ$/0.1 m} & \multirow{3}{*}{1 cam} & BEVFormer & 0.4648 & 0.3243 & \textbf{0.6871} & 0.2777 & 0.3866 & 0.4275 & 0.1951 \\
&  & CAPE & 0.4730 & \textbf{0.3623} & 0.6997 & \textbf{0.2649} & 0.4032 & 0.5217 & \textbf{0.1917} \\
&  & \textbf{NCGR} & \textbf{0.4857} & 0.3593 & 0.7332 & 0.2711 & \textbf{0.3781} & \textbf{0.3559} & 0.2010 \\
\cmidrule(l){2-10}
& \multirow{3}{*}{2 cam} & BEVFormer & 0.4442 & 0.3119 & \textbf{0.7297} & 0.2823 & 0.4339 & 0.4613 & 0.2107 \\
&  & CAPE & 0.4754 & 0.3601 & 0.7493 & \textbf{0.2666} & 0.4124 & 0.4192 & \textbf{0.1995} \\
&  & \textbf{NCGR} & \textbf{0.4823} & \textbf{0.3657} & 0.7548 & 0.2759 & \textbf{0.4050} & \textbf{0.3676} & 0.2021 \\
\cmidrule(l){2-10}
& \multirow{3}{*}{3 cam} & BEVFormer & 0.3757 & 0.2244 & 0.8370 & 0.2864 & 0.4631 & 0.5702 & 0.2080 \\
&  & CAPE & 0.4093 & 0.2843 & 0.8421 & \textbf{0.2667} & \textbf{0.4252} & 0.5886 & 0.2055 \\
&  & \textbf{NCGR} & \textbf{0.4553} & \textbf{0.3286} & \textbf{0.7989} & 0.2782 & 0.4284 & \textbf{0.3924} & \textbf{0.1915} \\
\cmidrule(l){2-10}
& \multirow{3}{*}{4 cam} & BEVFormer & 0.3735 & 0.2263 & 0.8387 & 0.2913 & 0.4833 & 0.5426 & 0.2402 \\
&  & CAPE & 0.4217 & 0.2966 & 0.8575 & \textbf{0.2698} & \textbf{0.4278} & 0.4911 & 0.2194 \\
&  & \textbf{NCGR} & \textbf{0.4411} & \textbf{0.3151} & \textbf{0.8349} & 0.2801 & 0.4583 & \textbf{0.3859} & \textbf{0.2049} \\
\cmidrule(l){2-10}
& \multirow{3}{*}{5 cam} & BEVFormer & 0.3016 & 0.1491 & 0.9857 & 0.2994 & 0.5344 & 0.6646 & 0.2450 \\
&  & CAPE & 0.3564 & 0.2337 & 0.9674 & \textbf{0.2697} & 0.4581 & 0.6798 & 0.2292 \\
&  & \textbf{NCGR} & \textbf{0.4092} & \textbf{0.2711} & \textbf{0.9072} & 0.2844 & \textbf{0.4570} & \textbf{0.4126} & \textbf{0.2025} \\
\bottomrule
\end{tabular}
\endgroup
\caption{Dynamic perturbation results across rotation bounds with a fixed translation bound of 0.1 m.}
\label{tab:app_dynamic_all}
\end{table}

\begin{table}[H]
\centering
\begingroup
\small
\renewcommand{\arraystretch}{0.65}
\setlength{\tabcolsep}{10pt}
\begin{tabular}{@{}cllccccccc@{}}
\toprule
Bound & \#Cam & Method & NDS$\uparrow$ & mAP$\uparrow$ & mATE$\downarrow$ & mASE$\downarrow$ & mAOE$\downarrow$ & mAVE$\downarrow$ & mAAE$\downarrow$ \\
\midrule
\multirow{15}{*}{$3^\circ$/0.1 m} & \multirow{3}{*}{1 cam} & BEVFormer & 0.4809 & 0.3640 & 0.7468 & 0.2767 & \textbf{0.3743} & 0.4147 & 0.1982 \\
&  & CAPE & \textbf{0.5026} & \textbf{0.3889} & \textbf{0.7448} & \textbf{0.2642} & 0.3965 & \textbf{0.3286} & \textbf{0.1845} \\
&  & \textbf{NCGR} & 0.4886 & 0.3688 & 0.7517 & 0.2721 & 0.3803 & 0.3490 & 0.2044 \\
\cmidrule(l){2-10}
& \multirow{3}{*}{2 cam} & BEVFormer & 0.4812 & 0.3675 & \textbf{0.7446} & 0.2785 & \textbf{0.3781} & 0.4181 & 0.2057 \\
&  & CAPE & \textbf{0.5057} & \textbf{0.3973} & 0.7504 & \textbf{0.2636} & 0.4004 & \textbf{0.3331} & \textbf{0.1820} \\
&  & \textbf{NCGR} & 0.4890 & 0.3739 & 0.7504 & 0.2756 & 0.3937 & 0.3563 & 0.2038 \\
\cmidrule(l){2-10}
& \multirow{3}{*}{3 cam} & BEVFormer & 0.4480 & 0.3238 & 0.8560 & 0.2789 & \textbf{0.3790} & 0.4219 & 0.2035 \\
&  & CAPE & \textbf{0.4763} & \textbf{0.3596} & 0.8519 & \textbf{0.2621} & 0.4031 & \textbf{0.3350} & \textbf{0.1825} \\
&  & \textbf{NCGR} & 0.4662 & 0.3419 & \textbf{0.8176} & 0.2765 & 0.3927 & 0.3572 & 0.2034 \\
\cmidrule(l){2-10}
& \multirow{3}{*}{4 cam} & BEVFormer & 0.4450 & 0.3252 & 0.8677 & 0.2820 & \textbf{0.3896} & 0.4260 & 0.2104 \\
&  & CAPE & \textbf{0.4854} & \textbf{0.3716} & \textbf{0.8227} & \textbf{0.2635} & 0.3980 & \textbf{0.3359} & \textbf{0.1833} \\
&  & \textbf{NCGR} & 0.4604 & 0.3390 & 0.8370 & 0.2774 & 0.4084 & 0.3628 & 0.2054 \\
\cmidrule(l){2-10}
& \multirow{3}{*}{5 cam} & BEVFormer & 0.4177 & 0.2924 & 0.9874 & 0.2820 & \textbf{0.3862} & 0.4237 & 0.2060 \\
&  & CAPE & \textbf{0.4574} & \textbf{0.3382} & \textbf{0.9206} & \textbf{0.2630} & 0.4027 & \textbf{0.3463} & \textbf{0.1848} \\
&  & \textbf{NCGR} & 0.4364 & 0.3082 & 0.9266 & 0.2792 & 0.4045 & 0.3615 & 0.2051 \\
\midrule
\multirow{15}{*}{$6^\circ$/0.1 m} & \multirow{3}{*}{1 cam} & BEVFormer & 0.4584 & 0.3087 & \textbf{0.6770} & 0.2804 & \textbf{0.3852} & 0.4173 & 0.2000 \\
&  & CAPE & 0.4733 & 0.3313 & 0.7103 & \textbf{0.2642} & 0.4020 & \textbf{0.3575} & \textbf{0.1894} \\
&  & \textbf{NCGR} & \textbf{0.4775} & \textbf{0.3498} & 0.7431 & 0.2740 & 0.3932 & 0.3615 & 0.2023 \\
\cmidrule(l){2-10}
& \multirow{3}{*}{2 cam} & BEVFormer & 0.4456 & 0.3117 & 0.7436 & 0.2809 & 0.4231 & 0.4464 & 0.2083 \\
&  & CAPE & 0.4687 & 0.3384 & 0.7747 & \textbf{0.2665} & \textbf{0.4129} & \textbf{0.3615} & \textbf{0.1895} \\
&  & \textbf{NCGR} & \textbf{0.4791} & \textbf{0.3600} & \textbf{0.7420} & 0.2747 & 0.4206 & 0.3683 & 0.2040 \\
\cmidrule(l){2-10}
& \multirow{3}{*}{3 cam} & BEVFormer & 0.3810 & 0.2217 & 0.8787 & 0.2838 & 0.4490 & 0.4792 & 0.2076 \\
&  & CAPE & 0.4187 & 0.2689 & 0.9072 & \textbf{0.2649} & \textbf{0.4330} & \textbf{0.3746} & \textbf{0.1782} \\
&  & \textbf{NCGR} & \textbf{0.4464} & \textbf{0.3109} & \textbf{0.7924} & 0.2776 & 0.4369 & 0.3893 & 0.1947 \\
\cmidrule(l){2-10}
& \multirow{3}{*}{4 cam} & BEVFormer & 0.3757 & 0.2244 & 0.9024 & 0.2908 & 0.4621 & 0.4794 & 0.2300 \\
&  & CAPE & 0.4279 & 0.2889 & 0.9010 & \textbf{0.2699} & \textbf{0.4266} & \textbf{0.3700} & \textbf{0.1986} \\
&  & \textbf{NCGR} & \textbf{0.4321} & \textbf{0.3007} & \textbf{0.8437} & 0.2805 & 0.4615 & 0.3876 & 0.2089 \\
\cmidrule(l){2-10}
& \multirow{3}{*}{5 cam} & BEVFormer & 0.3283 & 0.1626 & 1.0580 & 0.2937 & 0.4936 & 0.5124 & 0.2301 \\
&  & CAPE & 0.3821 & 0.2279 & 1.0370 & \textbf{0.2697} & \textbf{0.4565} & \textbf{0.3988} & \textbf{0.1935} \\
&  & \textbf{NCGR} & \textbf{0.3918} & \textbf{0.2470} & \textbf{0.9309} & 0.2849 & 0.4916 & 0.4103 & 0.1988 \\
\midrule
\multirow{15}{*}{$9^\circ$/0.1 m} & \multirow{3}{*}{1 cam} & BEVFormer & 0.4516 & 0.2900 & \textbf{0.6616} & 0.2800 & \textbf{0.3878} & 0.4056 & 0.1994 \\
&  & CAPE & 0.4570 & 0.2989 & 0.6791 & \textbf{0.2654} & 0.4020 & 0.3873 & \textbf{0.1913} \\
&  & \textbf{NCGR} & \textbf{0.4760} & \textbf{0.3488} & 0.7347 & 0.2751 & 0.4006 & \textbf{0.3676} & 0.2059 \\
\cmidrule(l){2-10}
& \multirow{3}{*}{2 cam} & BEVFormer & 0.4257 & 0.2766 & \textbf{0.7234} & 0.2852 & 0.4362 & 0.4678 & 0.2137 \\
&  & CAPE & 0.4458 & 0.3010 & 0.7673 & \textbf{0.2675} & 0.4322 & 0.3867 & \textbf{0.1936} \\
&  & \textbf{NCGR} & \textbf{0.4795} & \textbf{0.3616} & 0.7433 & 0.2741 & \textbf{0.4202} & \textbf{0.3681} & 0.2073 \\
\cmidrule(l){2-10}
& \multirow{3}{*}{3 cam} & BEVFormer & 0.3452 & 0.1548 & 0.8250 & 0.2901 & 0.4671 & 0.5249 & 0.2148 \\
&  & CAPE & 0.3814 & 0.2105 & 0.8929 & \textbf{0.2672} & 0.4693 & 0.4236 & \textbf{0.1855} \\
&  & \textbf{NCGR} & \textbf{0.4464} & \textbf{0.3104} & \textbf{0.7848} & 0.2774 & \textbf{0.4356} & \textbf{0.3971} & 0.1926 \\
\cmidrule(l){2-10}
& \multirow{3}{*}{4 cam} & BEVFormer & 0.3408 & 0.1674 & \textbf{0.8363} & 0.2955 & 0.5061 & 0.5440 & 0.2468 \\
&  & CAPE & 0.3925 & 0.2348 & 0.8971 & \textbf{0.2743} & \textbf{0.4643} & 0.4025 & 0.2109 \\
&  & \textbf{NCGR} & \textbf{0.4220} & \textbf{0.2876} & 0.8534 & 0.2806 & 0.4809 & \textbf{0.3940} & \textbf{0.2086} \\
\cmidrule(l){2-10}
& \multirow{3}{*}{5 cam} & BEVFormer & 0.2536 & 0.0767 & 0.9682 & 0.3488 & \textbf{0.4986} & 0.7070 & 0.3244 \\
&  & CAPE & 0.3361 & 0.1604 & 1.0616 & \textbf{0.2745} & 0.5023 & 0.4545 & 0.2090 \\
&  & \textbf{NCGR} & \textbf{0.3749} & \textbf{0.2218} & \textbf{0.9357} & 0.2853 & 0.5093 & \textbf{0.4351} & \textbf{0.1951} \\
\midrule
\multirow{15}{*}{$12^\circ$/0.1 m} & \multirow{3}{*}{1 cam} & BEVFormer & 0.4479 & 0.2815 & 0.6606 & 0.2829 & \textbf{0.3851} & 0.4006 & 0.1990 \\
&  & CAPE & 0.4492 & 0.2820 & \textbf{0.6597} & \textbf{0.2658} & 0.3988 & 0.4019 & \textbf{0.1915} \\
&  & \textbf{NCGR} & \textbf{0.4727} & \textbf{0.3439} & 0.7374 & 0.2751 & 0.3959 & \textbf{0.3792} & 0.2041 \\
\cmidrule(l){2-10}
& \multirow{3}{*}{2 cam} & BEVFormer & 0.4135 & 0.2557 & \textbf{0.7169} & 0.2879 & 0.4420 & 0.4789 & 0.2173 \\
&  & CAPE & 0.4319 & 0.2775 & 0.7565 & \textbf{0.2676} & 0.4445 & 0.4029 & \textbf{0.1965} \\
&  & \textbf{NCGR} & \textbf{0.4752} & \textbf{0.3552} & 0.7421 & 0.2762 & \textbf{0.4233} & \textbf{0.3739} & 0.2087 \\
\cmidrule(l){2-10}
& \multirow{3}{*}{3 cam} & BEVFormer & 0.3215 & 0.1180 & 0.8106 & 0.2970 & 0.4880 & 0.5598 & 0.2199 \\
&  & CAPE & 0.3591 & 0.1753 & 0.8759 & \textbf{0.2690} & 0.4951 & 0.4578 & \textbf{0.1879} \\
&  & \textbf{NCGR} & \textbf{0.4355} & \textbf{0.2949} & \textbf{0.7857} & 0.2797 & \textbf{0.4541} & \textbf{0.4031} & 0.1964 \\
\cmidrule(l){2-10}
& \multirow{3}{*}{4 cam} & BEVFormer & 0.3231 & 0.1409 & \textbf{0.8129} & 0.2962 & 0.5094 & 0.5899 & 0.2653 \\
&  & CAPE & 0.3713 & 0.2021 & 0.8910 & \textbf{0.2747} & \textbf{0.4887} & 0.4261 & 0.2176 \\
&  & \textbf{NCGR} & \textbf{0.4040} & \textbf{0.2621} & 0.8592 & 0.2819 & 0.5102 & \textbf{0.4043} & \textbf{0.2153} \\
\cmidrule(l){2-10}
& \multirow{3}{*}{5 cam} & BEVFormer & 0.2296 & 0.0423 & \textbf{0.9313} & 0.3529 & \textbf{0.5292} & 0.7697 & 0.3322 \\
&  & CAPE & 0.3051 & 0.1191 & 1.0559 & \textbf{0.2811} & 0.5533 & 0.4925 & 0.2182 \\
&  & \textbf{NCGR} & \textbf{0.3540} & \textbf{0.1934} & 0.9547 & 0.2858 & 0.5347 & \textbf{0.4442} & \textbf{0.2078} \\
\bottomrule
\end{tabular}
\endgroup
\caption{Static perturbation results across rotation bounds with a fixed translation bound of 0.1 m.}
\label{tab:app_static_all}
\end{table}

\begin{table}[H]
\ContinuedFloat
\centering
\begingroup
\small
\renewcommand{\arraystretch}{0.86}
\setlength{\tabcolsep}{10pt}
\begin{tabular}{@{}cllccccccc@{}}
\toprule
Bound & \#Cam & Method & NDS$\uparrow$ & mAP$\uparrow$ & mATE$\downarrow$ & mASE$\downarrow$ & mAOE$\downarrow$ & mAVE$\downarrow$ & mAAE$\downarrow$ \\
\midrule
\multirow{15}{*}{$15^\circ$/0.1 m} & \multirow{3}{*}{1 cam} & BEVFormer & 0.4459 & 0.2752 & 0.6599 & 0.2827 & \textbf{0.3778} & 0.3988 & 0.1979 \\
&  & CAPE & 0.4456 & 0.2766 & \textbf{0.6542} & \textbf{0.2665} & 0.3999 & 0.4159 & \textbf{0.1906} \\
&  & \textbf{NCGR} & \textbf{0.4666} & \textbf{0.3325} & 0.7331 & 0.2733 & 0.4029 & \textbf{0.3849} & 0.2026 \\
\cmidrule(l){2-10}
& \multirow{3}{*}{2 cam} & BEVFormer & 0.4042 & 0.2386 & \textbf{0.7187} & 0.2900 & 0.4412 & 0.4853 & 0.2155 \\
&  & CAPE & 0.4226 & 0.2626 & 0.7443 & \textbf{0.2688} & 0.4586 & 0.4183 & \textbf{0.1966} \\
&  & \textbf{NCGR} & \textbf{0.4681} & \textbf{0.3417} & 0.7485 & 0.2770 & \textbf{0.4240} & \textbf{0.3712} & 0.2073 \\
\cmidrule(l){2-10}
& \multirow{3}{*}{3 cam} & BEVFormer & 0.3075 & 0.0953 & 0.8160 & 0.3023 & 0.4884 & 0.5785 & 0.2159 \\
&  & CAPE & 0.3439 & 0.1537 & 0.8569 & \textbf{0.2709} & 0.5207 & 0.4938 & \textbf{0.1869} \\
&  & \textbf{NCGR} & \textbf{0.4208} & \textbf{0.2669} & \textbf{0.7983} & 0.2805 & \textbf{0.4568} & \textbf{0.4003} & 0.1903 \\
\cmidrule(l){2-10}
& \multirow{3}{*}{4 cam} & BEVFormer & 0.3162 & 0.1247 & \textbf{0.8181} & 0.3011 & \textbf{0.5023} & 0.5906 & 0.2492 \\
&  & CAPE & 0.3569 & 0.1818 & 0.8800 & \textbf{0.2762} & 0.5252 & 0.4394 & \textbf{0.2186} \\
&  & \textbf{NCGR} & \textbf{0.3860} & \textbf{0.2309} & 0.8545 & 0.2815 & 0.5251 & \textbf{0.4076} & 0.2258 \\
\cmidrule(l){2-10}
& \multirow{3}{*}{5 cam} & BEVFormer & 0.2412 & 0.0314 & \textbf{0.9399} & 0.3256 & \textbf{0.5364} & 0.6884 & 0.2545 \\
&  & CAPE & 0.2868 & 0.0942 & 1.0457 & 0.2850 & 0.5944 & 0.5061 & 0.2169 \\
&  & \textbf{NCGR} & \textbf{0.3373} & \textbf{0.1587} & 0.9576 & \textbf{0.2846} & 0.5386 & \textbf{0.4394} & \textbf{0.2005} \\
\bottomrule
\end{tabular}
\endgroup
\caption{Static perturbation results across rotation bounds with a fixed translation bound of 0.1 m. (continued)}
\begin{minipage}{0.96\textwidth}
\subsection{Class-Wise Robustness under Five-Camera Perturbations}
Table~\ref{tab:app_classwise_five_camera} reports class-wise mean distance AP
for the predefined five-camera setting at $15^\circ$/0.1 m. NCGR attains the
highest value for each of the ten nuScenes classes under both dynamic and
static perturbations. The class-wise breakdown shows that the aggregate gains
in NDS and mAP are distributed across object categories rather than being
driven by a small subset of classes.

\centering
\begingroup
\small
\setlength{\tabcolsep}{6.5pt}
\begin{tabular}{@{}lcccccc@{}}
\toprule
& \multicolumn{3}{c}{Dynamic, 5 cam} &
  \multicolumn{3}{c}{Static, 5 cam} \\
\cmidrule(lr){2-4}\cmidrule(l){5-7}
Class & BEVFormer & CAPE & NCGR & BEVFormer & CAPE & NCGR \\
\midrule
Car                  & 0.1557 & 0.2390 & \textbf{0.3482} & 0.0550 & 0.1057 & \textbf{0.1959} \\
Truck                & 0.0666 & 0.1299 & \textbf{0.1847} & 0.0131 & 0.0373 & \textbf{0.0954} \\
Bus                  & 0.1507 & 0.2083 & \textbf{0.2793} & 0.0887 & 0.1128 & \textbf{0.1855} \\
Trailer              & 0.0056 & 0.1115 & \textbf{0.1133} & 0.0000 & 0.0379 & \textbf{0.0413} \\
Construction vehicle & 0.0049 & 0.0392 & \textbf{0.0488} & 0.0000 & 0.0058 & \textbf{0.0172} \\
Pedestrian           & 0.1444 & 0.2410 & \textbf{0.3127} & 0.0270 & 0.0901 & \textbf{0.1790} \\
Motorcycle           & 0.1277 & 0.2210 & \textbf{0.2743} & 0.0402 & 0.1089 & \textbf{0.1784} \\
Bicycle              & 0.0746 & 0.1703 & \textbf{0.2285} & 0.0087 & 0.0587 & \textbf{0.1102} \\
Traffic cone         & 0.1895 & 0.3355 & \textbf{0.3847} & 0.0283 & 0.1920 & \textbf{0.2750} \\
Barrier              & 0.2141 & 0.3195 & \textbf{0.3848} & 0.0531 & 0.1923 & \textbf{0.3096} \\
\bottomrule
\end{tabular}
\endgroup
\captionof{table}{Class-wise mean distance AP under the predefined five-camera
$15^\circ$/0.1 m perturbation setting. Dynamic perturbations are resampled per
validation sample, whereas the static realization is fixed throughout
validation. The best result in each setting and class is shown in bold.}
\label{tab:app_classwise_five_camera}
\end{minipage}
\end{table}

\section{Mechanism and Rectification Evidence}
\label{app:mechanism_qualitative}

\subsection{Controlled Rectification-Path Intervention}
We isolate the inference-time contribution of the rectification path using the
same trained checkpoint and the predefined five-camera dynamic setting at
$15^\circ$/0.1 m. In the offset-disabled intervention, the effective offset
$\boldsymbol{\delta}_{j,i}^{\star}$ in
Eq.~\eqref{eq:app_ncgr_execution} is set to zero while all remaining model
components are unchanged. An independent gate-closed intervention sets the
rectification gate to its closed endpoint. Both interventions produce closely
matched reductions in NDS and mAP, linking the full-model gain to the execution of
the gated reference-point correction.

\begin{table}[H]
\centering
\begingroup
\small
\setlength{\tabcolsep}{5.5pt}
\begin{tabular}{@{}lccc@{}}
\toprule
Variant & NDS$\uparrow$ & mAP$\uparrow$ & mAVE$\downarrow$ \\
\midrule
Full NCGR       & \textbf{0.3969} & \textbf{0.2559} & \textbf{0.4229} \\
Offset disabled & 0.3434 & 0.1892 & 0.4834 \\
Gate closed     & 0.3445 & 0.1891 & 0.5042 \\
\bottomrule
\end{tabular}
\endgroup
\caption{Controlled inference-time interventions under the five-camera dynamic
$15^\circ$/0.1 m perturbation setting. All rows use the same trained NCGR
checkpoint.}
\label{tab:app_rectification_intervention}
\end{table}

\begin{figure}[H]
\raggedright
\subsection{Qualitative 3D Detection Results}
\centering
\includegraphics[width=0.90\textwidth]{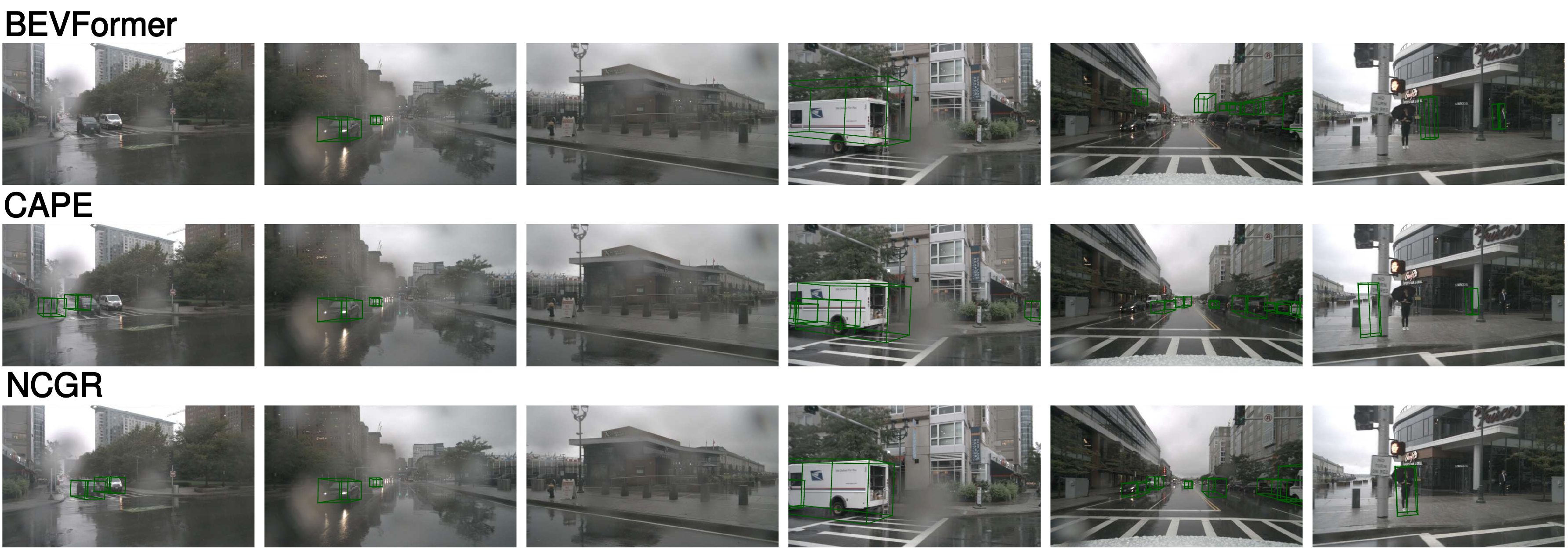}
\caption{Qualitative 3D detection results under multi-camera extrinsic
perturbations for BEVFormer, CAPE, and NCGR.}
\label{fig:supp_qualitative_detection}

\bigskip
\raggedright
\subsection{Reference-Point Rectification Visualization}
\centering
\includegraphics[width=0.88\textwidth]{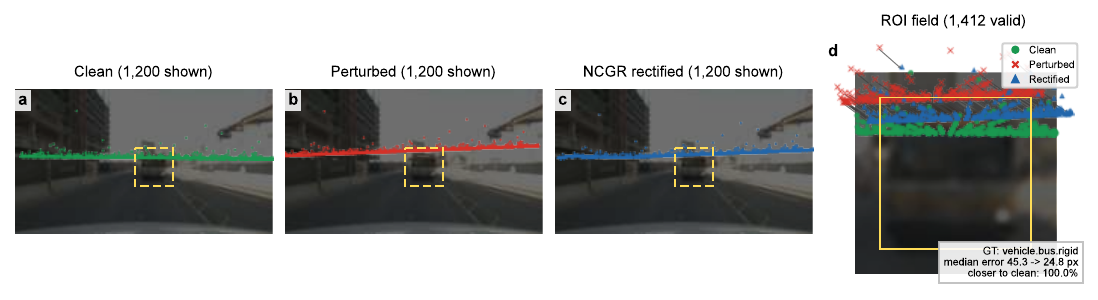}
\caption{Visualization of reference-point rectification. Clean, perturbed, and
rectified image-plane projections are shown for representative BEV reference
points. Extrinsic perturbations displace the original projection anchors,
whereas NCGR shifts them toward the corresponding clean locations before
native deformable sampling. Panels~a--c display the same deterministic subset
of 1,200 valid full-image reference points for visual clarity. Panel~d enlarges
the yellow-boxed region of interest (ROI), which is a local image area around
the target object. For readability, it displays deterministic subsets of 320
points for each projection state and 140 rectification arrows. Its title and
inset statistics are computed over all 1,412 valid displaced reference points
inside the ROI. The median Euclidean image-plane distance to the corresponding
clean projection decreases from 45.3 to 24.8 pixels, and all 1,412 points move
closer to their clean projections. In the inset, GT denotes ground truth.}
\label{fig:supp_reference_rectification}
\end{figure}

\subsection{Query- and Camera-Conditioned Rectification Fields}
\begin{figure}[H]
\centering
\includegraphics[width=0.84\textwidth]{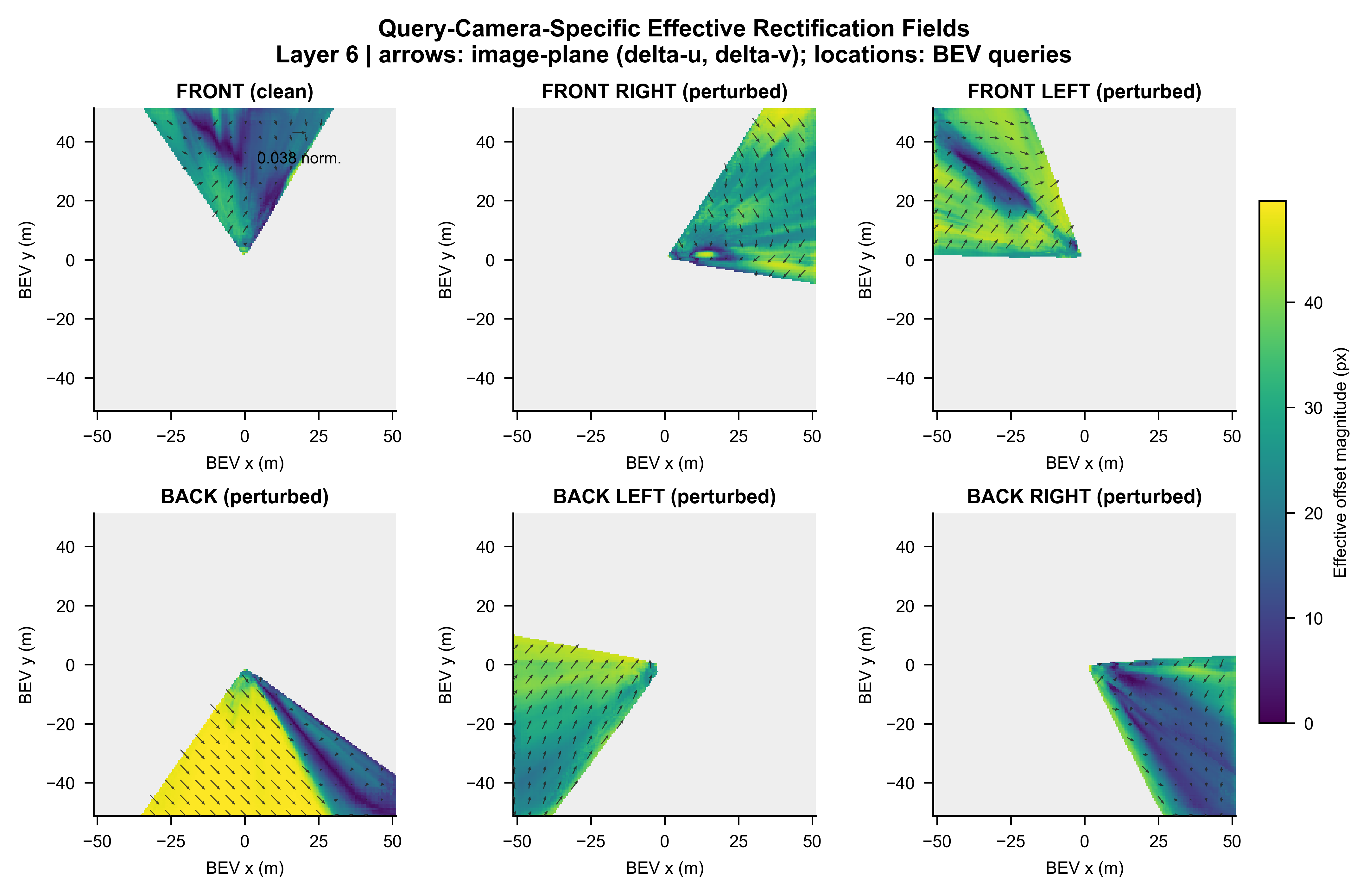}
\caption{Query--camera-specific effective rectification fields at SCA
Layer~6. The six panels show the same validation frame for
\texttt{CAM\_FRONT}, \texttt{CAM\_FRONT\_RIGHT},
\texttt{CAM\_FRONT\_LEFT}, \texttt{CAM\_BACK},
\texttt{CAM\_BACK\_LEFT}, and \texttt{CAM\_BACK\_RIGHT}, respectively.
\texttt{CAM\_FRONT} is clean, and the remaining cameras are perturbed. Arrows
are anchored at BEV-query locations but represent image-plane offsets. Their
directions and relative lengths encode offset direction and magnitude.
Background color gives the corresponding magnitude in padded-image pixels, and
gray regions contain no valid query--camera projection.}
\label{fig:supp_query_camera_fields}
\end{figure}

To determine whether NCGR predicts different corrections for different cameras
and spatial locations, Fig.~\ref{fig:supp_query_camera_fields} visualizes all
six camera streams from the same validation scene at SCA Layer~6.
\texttt{CAM\_FRONT} remains clean, whereas the other five cameras receive
extrinsic perturbations.

The arrows indicate the predicted image-plane rectification directions, with
their lengths scaled for relative visual comparison. The background color
quantifies the corresponding effective offset magnitude in padded-image
pixels. Gray regions indicate BEV queries without valid projections in the
corresponding camera. The scale annotation ``0.038 norm.'' marks a reference
arrow whose Euclidean magnitude is 0.038 in the normalized image-plane
coordinates used by SCA.

\FloatBarrier
The six cameras exhibit different offset directions and spatial structures.
This shows that the model does not apply the same translation to all cameras
and spatial locations. Although \texttt{CAM\_FRONT} is unperturbed, its
Layer~6 BEV queries already contain multi-camera information aggregated by the
preceding layers.
In jointly visible regions, these queries may carry inconsistencies introduced
by adjacent perturbed cameras. Therefore, nonzero corrections in the clean
camera can arise from the shared multi-view context and should not be
interpreted as evidence that the clean camera has been classified as
perturbed.

\subsection{Layer-Wise Projection-Error Analysis}
\begin{figure}[H]
\centering
\includegraphics[width=0.84\textwidth]{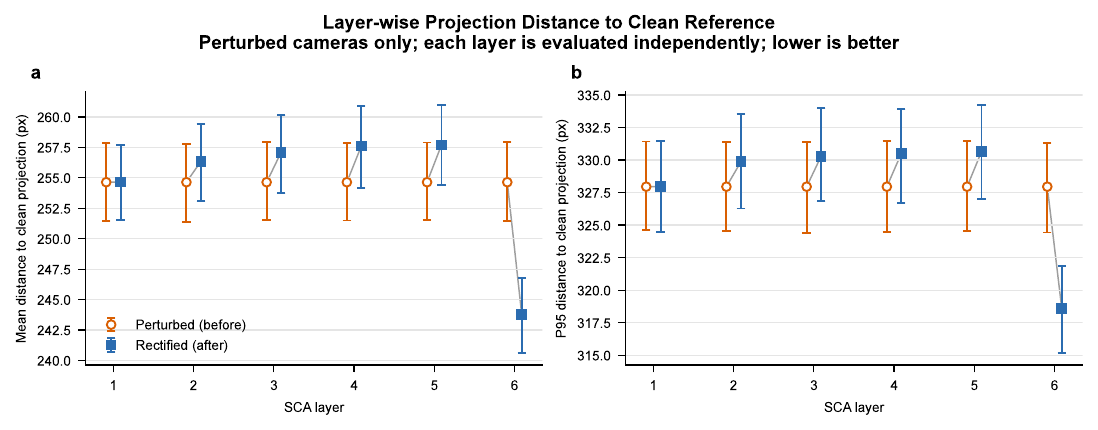}
\caption{Layer-wise projection distance to the clean reference under
five-camera dynamic perturbations, with each SCA layer evaluated independently.
Orange open circles show the error before rectification, and blue squares show
the error after applying that layer's effective offset; lower values are closer
to the clean references. Panel~a reports the mean, and Panel~b reports P95, the
95th-percentile error. In each frame, the metric is computed separately for the
five perturbed cameras and then averaged across them. The plotted values average
these summaries over 600 validation frames; error bars show 95\% bootstrap
confidence intervals over the validation frames.}
\label{fig:supp_layerwise_projection_error}
\end{figure}

Figure~\ref{fig:supp_layerwise_projection_error} examines which SCA layer
predicts offsets that move perturbed projections closer to their clean
references. For every valid projection, we measure the distance to its clean
location before and after applying the effective offset from each layer.
We denote these pre- and post-rectification distances by
$e_{\mathrm{pert}}$ and $e_{\mathrm{rect}}$, respectively. Although BEV
features are propagated sequentially through the encoder, each layer-specific
offset is evaluated separately on the same original perturbed projection and
is not accumulated with offsets from the preceding layers. The analysis uses
600 validation frames under the five-camera dynamic $15^\circ$/0.1\,m
perturbation setting.

\begin{figure}[H]
\centering
\includegraphics[width=0.80\textwidth]{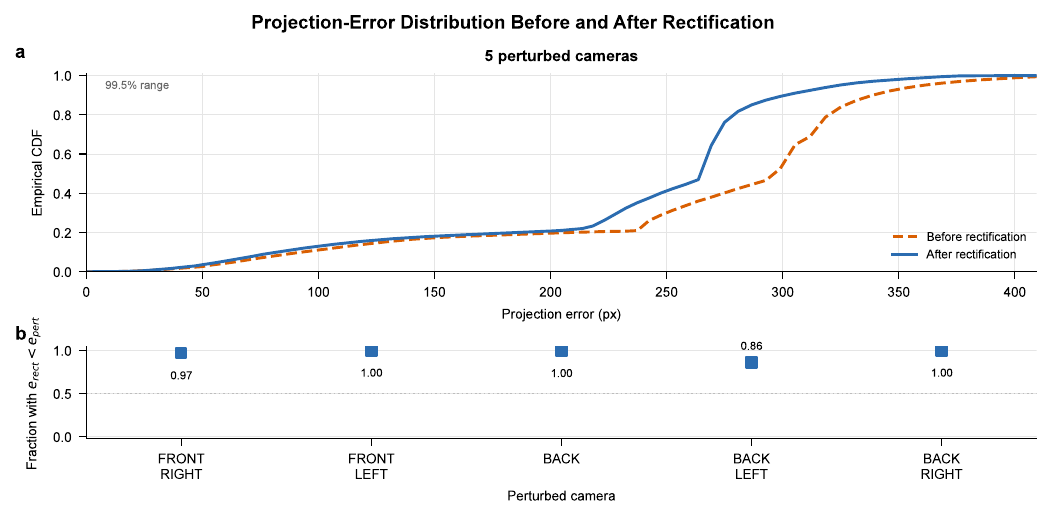}
\caption{Projection-error distributions before and after Layer~6
rectification in a selected dynamic training frame. Panel~a shows the
empirical cumulative distribution functions (CDFs) of valid errors pooled
across the five perturbed cameras. At a fixed threshold, a blue curve above the
orange curve means that more points fall below that threshold after
rectification; the displayed range covers 99.5\% of the pooled projections.
Panel~b reports the per-camera fraction satisfying
$e_{\mathrm{rect}}<e_{\mathrm{pert}}$; the dotted line marks 0.5.}
\label{fig:supp_projection_error_distribution}
\end{figure}

\FloatBarrier
Layers~1--5 yield little reduction in projection error, and several layers
produce small increases. Layer~6 is the only layer that clearly reduces both
statistics. The mean error decreases from 254.6 to 243.8 pixels, corresponding
to a 4.3\% reduction.
The P95 error decreases from 328.0 to 318.6 pixels, corresponding to a 2.8\%
reduction. Under this experimental setting, Layer~6 predicts the most
geometrically effective rectification. Figure~\ref{fig:supp_projection_error_distribution}
shows how the projection points change within a selected dynamic training
frame.

\FloatBarrier
The blue curve lies above the orange curve through most of the displayed range,
showing that more projection points fall below the same error thresholds after
rectification. The mean error pooled across the five perturbed cameras
decreases from 260.2 to 231.8 pixels, and 96.4\% of valid projection points
move closer to their clean references.
Panel~b reports the fraction of improved points for each perturbed camera. The
five fractions are 96.8\%, 100.0\%, 99.5\%, 86.1\%, and 100.0\%, respectively.

\end{document}